%% file: main.tex
\documentclass[sigconf,nonacm]{acmart}
\AtBeginDocument{%
  }

\usepackage{float}        
\usepackage{placeins}     
\usepackage{dblfloatfix}  
\usepackage{afterpage}
\usepackage{placeins}     
\usepackage{dblfloatfix}  
\usepackage[normalem]{ulem}

\setcopyright{acmlicensed}
\copyrightyear{2025}
\acmYear{2025}
\acmDOI{XXXXXXX.XXXXXXX}
\acmConference[ICAIF '25]{ACM International Conference on AI in Finance}{November 15-18, 2025}{Singapore}

\input{preamble}

\title{QuantEvolve: Automating Quantitative Strategy Discovery through Multi-Agent Evolutionary Framework}

\author{
  Junhyeog Yun\textsuperscript{*} \quad
  Hyoun Jun Lee\textsuperscript{*} \quad
  Insu Jeon\textsuperscript{\dag}
}

\affiliation{%
  \institution{AI Tech Lab, Qraft Technologies}
  \city{}
  \country{}
}
\email{{junhyeog.yun,   hyounjun.lee,   insu.jeon}@qraftec.com}

\thanks{\textsuperscript{*}Equal Contribution}
\thanks{\textsuperscript{\dag}Corresponding Author}

\begin{document}

\renewcommand{\shortauthors}{Junhyeog Yun, Hyounjun Lee, and Insu Jeon}

\begin{abstract}

Automating quantitative trading strategy development in dynamic markets is challenging, especially with increasing demand for personalized investment solutions.
Existing methods often fail to explore the vast strategy space while preserving the diversity essential for robust performance across changing market conditions.
We present QuantEvolve, an evolutionary framework that combines quality-diversity optimization with hypothesis-driven strategy generation. QuantEvolve employs a feature map aligned with investor preferences—such as strategy type, risk profile, turnover, and return characteristics—to maintain a diverse set of effective strategies. It also integrates a hypothesis-driven multi-agent system to systematically explore the strategy space through iterative generation and evaluation.
This approach produces diverse, sophisticated strategies that adapt to both market regime shifts and individual investment needs. Empirical results show that QuantEvolve outperforms conventional baselines, validating the effectiveness. We release a dataset of evolved strategies to support future research.

\end{abstract}


\maketitle

\begin{figure*}[!t]
  \centering
  \includegraphics[trim={0.3cm 0.3cm 0.3cm 0.3cm}, clip, width=\textwidth,keepaspectratio]{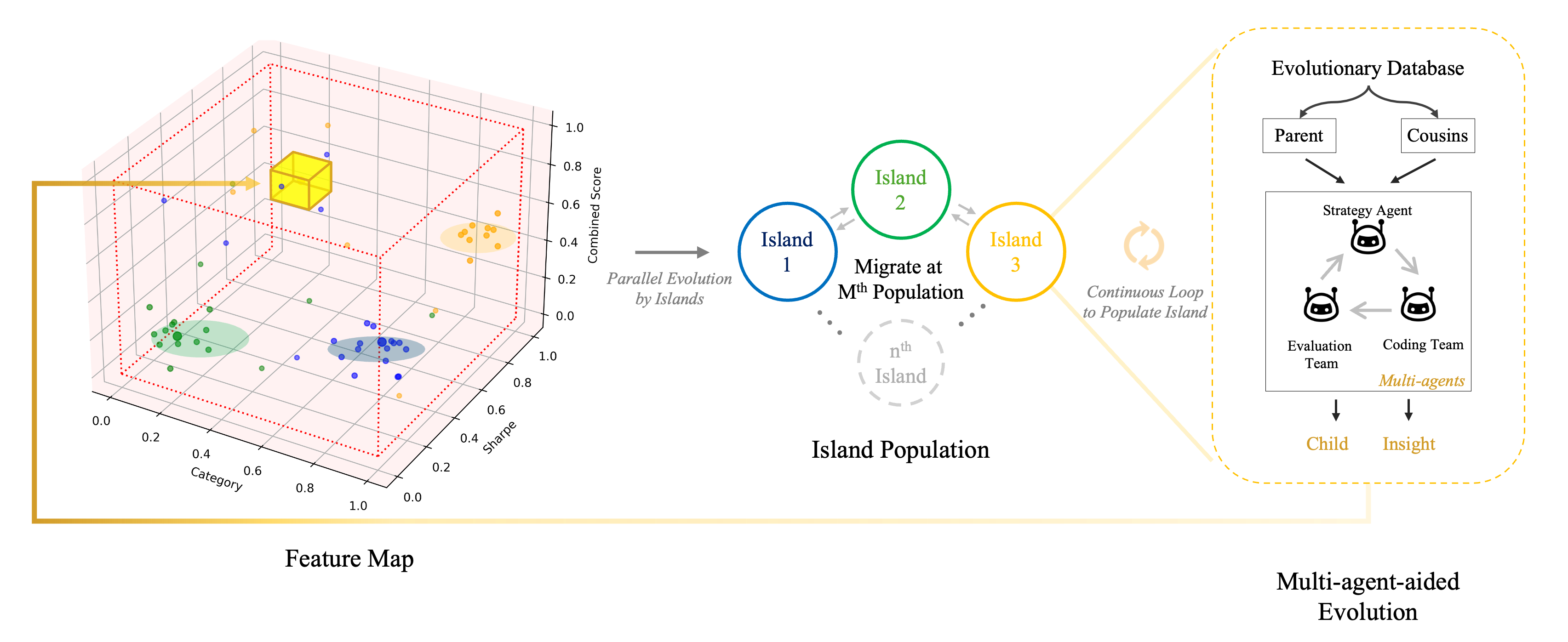}
  \caption{Architecture of the proposed QuantEvolve framework.}
  \label{fig:architecture-overview}
\end{figure*}

\input{sec/introduction}

\FloatBarrier

\section{Related Work}

\subsection{LLM Agents In Finance}
The emergence of LLMs and their derivative agentic architectures has generated transformative applications across multiple industries, with finance among the most prominent domains. In this context, LLM-based agents have enabled advances in portfolio construction~\cite{yu2024fincon,guo2025mass}, explainable investment products~\cite{liu2023fingpt,fatemi2024finvision}, and signal processing~\cite{yu2025finmem,zhang2024multimodal}, broadening the analytical toolkit available to practitioners. Frameworks such as AlphaAgents and FinRobot demonstrate how agentic systems can integrate heterogeneous financial data sources and uncover insights beyond traditional human-driven methods~\cite{zhao2025alphaagents,zhou2024finrobot}. Yet most existing efforts remain oriented toward narrow subtasks and rely heavily on semantic cues, making them difficult to backtest and limiting their robustness in real-world settings. These challenges motivate further exploration into how multi-agent systems can be designed to address the more demanding requirements of quantitative trading.

\subsection{LLM Agents in Quant Trading}
Subsequent research focuses on multi-agent frameworks that can be empirically tested with reusable evaluation methodologies. AlphaGPT and R\&D-Agent-Quant demonstrate how specialized agents can collaborate to generate and refine trading strategies through distributed processing and iterative improvement~\cite{yuan2024alpha,li2025r}. These systems primarily focus on alpha factor discovery, identifying individual predictive signals that require subsequent integration into complete trading systems. While such approaches show promise in controlled environments, they remain constrained by architectural limitations that restrict deployment to narrow factor mining tasks rather than comprehensive strategy development. Current implementations lack mechanisms for developing end-to-end trading systems that integrate multiple interacting factors with execution logic, position sizing, and risk management components. This persistent gap highlights the imperative to advance multi-agent orchestration toward architectures that can systematically generate complete trading strategies through scientific methodology, moving beyond individual factor discovery to maintain empirical rigor, ensure reproducibility, and deliver operational adaptability across diverse market regimes.

\subsection{Evolutionary LLM}
The limitations of traditional multi-agent frameworks have prompted exploration of evolutionary approaches that systematically navigate complex strategy spaces while maintaining behavioral diversity. Recent systems such as AI Scientist~\cite{lu2024ai} and AlphaEvolve~\cite{novikov2025alphaevolve} demonstrate how evolutionary algorithms can enhance multi-agent coordination by preventing mode collapse and promoting novel solution exploration. These approaches, inspired by Darwin-Gödel machines~\cite{zhang2025darwin} that combine evolutionary principles with formal reasoning, have shown promise in general AI domains but remain largely unexplored in quantitative trading. Building upon these insights, our work introduces an evolutionary multi-agent architecture specifically designed for quantitative strategy development, combining population-based diversity preservation with hypothesis-driven strategy generation to achieve robust performance across varying market conditions.

\input{sec/method}
\input{sec/experimental_setup}
\input{sec/result}

\section{Discussion}
\noindent \textbf{Robustness and Overfitting.}
This study focuses on applying evolutionary frameworks to quantitative finance; consequently, comprehensive robustness testing remains an area for improvement. As such, strategies generated by the framework may be susceptible to data snooping bias.
Future work will integrate more sophisticated and rigorous validation methodologies.

\noindent \textbf{Hypothesis Quality.} We have not formally validated whether generated hypotheses reflect established market theories or provide post-hoc rationalizations for data-mined patterns. Future work should incorporate external validation mechanisms—comparison against academic literature, expert review, or causal inference frameworks—to verify hypotheses represent meaningful insights.

\noindent \textbf{LLM Inference Cost.} Each evolutionary cycle requires 5-10 LLM inferences. This limits scalability for resource-constrained settings.

\noindent \textbf{Evaluation Scope.} Our evaluation uses six equities and two futures over limited date ranges with relatively simple baselines. While adequate for demonstrating comparative performance, these baselines do not represent state-of-the-art quantitative approaches. Future work should evaluate against sophisticated benchmarks and expand to larger asset universes with longer time horizons.

Fully automated quantitative research remains an open challenge. While QuantEvolve demonstrates the feasibility of an evolutionary framework for generating diverse strategies, substantial work is needed to ensure robustness, validate the quality of hypotheses, and bridge the research-to-deployment gap.

\section{Conclusion}
We present QuantEvolve, a multi-agent-aided evolutionary framework that systematically generates diverse, high-performing trading strategies. 
QuantEvolve employs a feature map that preserves population diversity across feature dimensions relevant to investor needs, and utilizes a multi-agent system to effectively explore a high-dimensional search space of trading strategies through hypothesis-driven reproduction.
By combining evolutionary computation's exploration with structured hypothesis-driven reasoning, QuantEvolve bridges the gap between broad discovery and deep, theoretically-grounded refinement.
Our empirical evaluation across equity and futures markets demonstrates that QuantEvolve produces diverse strategies with distinct behavioral characteristics, suggesting our framework can complement human expertise by exploring combinations and inefficiencies at scales infeasible for manual research.
By generating strategies across the full spectrum of risk profiles and trading philosophies, QuantEvolve enables personalized asset management adapting to individual constraints—a capability increasingly demanded yet underserved by existing platforms.

\section{Disclaimer}
This framework is intended for research purposes only. Users are solely responsible for validating all generated strategies and assessing their suitability for specific use cases. The framework's outputs do not reflect the opinions of Qraft Technologies. Qraft Technologies assumes no liability for any outcomes resulting from the use of this framework.

\bibliographystyle{ACM-Reference-Format}
\bibliography{bibliography}

\input{sec/appendix}

\end{document}

%% file: preamble.tex
\usepackage{tikz}
\usetikzlibrary{tikzmark,calc}

\usepackage{multicol}
\usepackage{algorithm}
\usepackage{algpseudocode}
\usepackage{algorithmicx}
\usepackage{amsmath}
\usepackage{amsthm}
\usepackage{mathtools}
\usepackage{subcaption}
\usepackage{pgfplots}
\usepackage{svg}
\usepackage{comment}
\usepackage{graphicx}
\usepackage{adjustbox}

\usepackage{dblfloatfix}

\usepackage{multirow}
\usepackage{soul} 

\usepackage{wrapfig}

\usepackage{enumitem}
\usepackage{tcolorbox}
\usepackage{listings}

\tcbuselibrary{breakable,skins}

\definecolor{lightred}{RGB}{249, 65, 68}
\definecolor{lightorange}{RGB}{248, 150, 30}
\definecolor{lightyellow}{RGB}{249, 199, 79}

\definecolor{lightblue}{RGB}{173, 216, 230}

\usepackage{pifont}

\tcbset{
    highlightstyle/.style={
        colback=cyan!20, 
        colframe=cyan!20, 
        arc=2pt, 
        boxrule=0pt, 
        boxsep=01pt, 
        left=1pt, right=1pt, top=1pt, bottom=1pt, 
        baseline=0pt, 
    }
}

\usepackage[capitalize]{cleveref}
\crefname{section}{Sec.}{Secs.}
\Crefname{section}{Section}{Sections}
\Crefname{table}{Table}{Tables}
\crefname{table}{Tab.}{Tabs.}

%% file: sec/introduction.tex
\section{Introduction}

As financial markets evolve, the demand for faster and more adaptive trading strategy development has become critical.
Traditional quantitative research relies on human researchers to design, test, and refine trading algorithms. 
This process involves analyzing historical data, identifying profitable patterns, and adapting to market conditions~\cite{treleaven2013algorithmic, bhuiyan2025deep, le2025quant, dakalbab2024ai}.
However, human researchers struggle to navigate diverse strategic frameworks due to inherent cognitive biases and limited attentional capacity~\cite{wang2023alpha, guo2024quant}.
These constraints become especially acute during rapid regime changes, as delays in the research process often lead to missed alpha opportunities~\cite{treleaven2013algorithmic, bhuiyan2025deep, le2025quant, dakalbab2024ai}.

Recent technological advances, particularly in large language models (LLMs) and multi-agent systems, have paved the way for sophisticated automation in quantitative finance. 
Researchers have begun deploying these systems throughout the investment pipeline, from discovering predictive alpha factors to executing real-time trading decisions, with the aim of supporting or extending traditional human-driven research~\cite{zhang2020autoalpha, yuan2024alpha}. 
Notable examples, such as R\&D-Agent-Quant and QuantAgent, leverage self-improving agent architectures to generate and refine trading strategies~\cite{wang2024quantagent, li2025r}. 
These studies demonstrate that LLMs can facilitate efficient development and implementation of trading algorithms.

Despite these advances, current LLM-based and agent-driven approaches still face important limitations in real-world applications.  
First, although they enhance the automation of strategy discovery, they often struggle to accommodate the diversity and complexity of modern financial environments~\cite{tu2010regime,bensaida2015frequency}.  
Most existing work focuses on optimizing isolated strategies or narrow objectives such as short-term returns, limiting their ability to account for the breadth of market conditions, trading styles, and risk regimes that coexist in practice.
Second, a major shortcoming of these automated systems lies in their inability to address the growing demand for personalized asset management.  
While scalable, such systems often overlook the fundamental heterogeneity in investor preferences.  
For example, direct indexing assets reached \$864.3 billion by December 2024, growing at a 22.4\% compound annual growth rate (CAGR).  
Robo-advisory platforms now manage \$1.4 trillion in assets, yet still serve only 5\% of U.S. investors, primarily due to limited personalization capabilities~\cite{cerulli2025directindexing, yahoo2025roboadvisor}.
These trends point to a critical gap: instead of relying on a single optimal strategy, future systems must dynamically generate and manage portfolios aligned with diverse investor preferences—each tailored to specific trading philosophies (e.g., momentum vs. mean-reversion), risk profiles, and investment horizons (e.g., high-frequency vs. low-turnover)—while maintaining adaptability to regime shifts.

To address these challenges, we present QuantEvolve, an evolutionary multi-agent framework, designed for the automated creation of trading strategies.
Leveraging recent advancements—such as AlphaEvolve and The AI Scientist—that demonstrate the success of evolutionary computation in tackling intricate scientific problems~\cite{lu2024ai, novikov2025alphaevolve}, we adapt these methodologies to quantitative finance.
QuantEvolve explores a diverse set of strategies using a feature map aligned with investor preferences.
It also utilizes a hypothesis-driven multi-agent system that facilitates a systematic exploration of the strategy search space through structured reasoning and iterative refinement during the evolutionary cycle. 
This integration enables QuantEvolve to generate a broad range of high-performance strategies adaptable to fluctuating market regimes while meeting the specific needs of personalized investment.

The primary contributions of this paper are as follows:
\begin{itemize}
    \item We introduce QuantEvolve, a multi-agent evolutionary framework that produces diverse trading strategies, adaptable to changing markets and personalized investment goals.
    
    \item We design a hypothesis-driven multi-agent system within an evolutionary framework that enables efficient exploration of a vast, high-dimensional strategy space, thereby facilitating systematic and in-depth research.
    
    \item We demonstrate that QuantEvolve produces strategies that outperform conventional baselines, validating its effectiveness and potential for broader use in automated quantitative strategy development.
    
    \item We release a detailed dataset of strategies generated by QuantEvolve, facilitating further research on automated quantitative finance and evolutionary strategy generation.
\end{itemize}

%% file: sec/method.tex
\section{Methodology}

\begin{figure*}[t]
  \centering
  \includegraphics[width=0.9\linewidth]{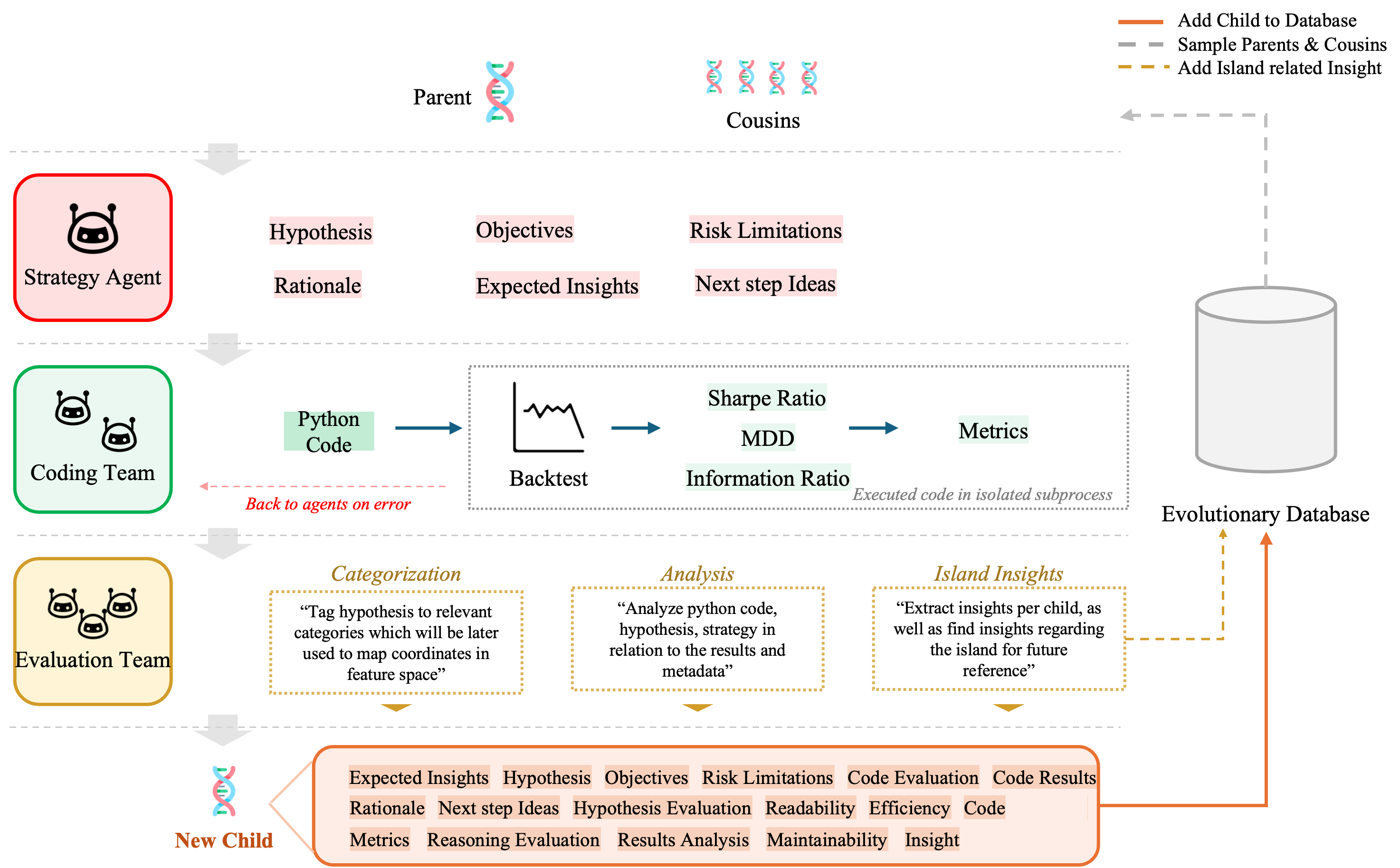}
  \Description{}
  \caption{Multi-agent evolution architecture.}
  \label{fig:multi-agent-arch}
\end{figure*}

We propose QuantEvolve, a multi-agent-aided evolutionary framework for generating diverse, high-performing trading strategies. QuantEvolve addresses two key challenges: maintaining population diversity across investor preferences and efficiently exploring the high-dimensional strategy search space \cite{mouret2015illuminating}.
To preserve diversity, we employ a feature map—a multi-dimensional archive that characterizes strategies by attributes aligned with investor needs (risk profile, trading frequency, return characteristics) and retains only the best performer in each behavioral niche. To enable effective search, we deploy a hypothesis-driven multi-agent system that systematically explores and refines promising strategy concepts through structured reasoning.
\Cref{fig:architecture-overview} illustrates the framework architecture. We describe the feature map design in \cref{sec:feature_map} and the multi-agent evolution process in \cref{sec:multi-agent}.

\begin{algorithm*}[t]
\caption{QuantEvolve}
\label{alg:quantevolve}
\begin{algorithmic}[1]
\Require Number of islands $N$, number of generations $G$, migration interval $M$, insight curation interval $K$, feature dimensions $\mathcal{D} = \{d_1, \ldots, d_D\}$, bin sizes per dimension $\mathcal{B} = \{B_1, \ldots, B_D\}$, exploitation-exploration balance $\alpha \in [0,1]$
\item[\textbf{Notation:}] $h$ (hypothesis), $c$ (code), $m$ (backtest results), $a$ (analysis)
\State \textbf{Initialization:}
\State Initialize feature map $\mathcal{F}$ with dimensions $\mathcal{D}$ and bin sizes $\mathcal{B}$
\State Initialize evolutionary database $\mathcal{DB}$ containing feature map and archive
\State Initialize $N$ islands $\{I_1, I_2, \ldots, I_N\}$ with seed strategies using \textsc{DataAgent}
\State Initialize insight repository $\mathcal{I} \leftarrow \emptyset$
\For{generation $g = 0$ to $G-1$}
    \For{each island $I_i$ where $i \in \{1, \ldots, N\}$}
        \State \textbf{// Parent and Cousin Sampling}
        \State $s_p \leftarrow$ \textsc{SampleParent}($I_i$, $\mathcal{DB}$, $\alpha$) \Comment{Sample parent strategy; \Cref{eq:parent_sampling}}
        \State $\mathcal{C} \leftarrow$ \textsc{SampleCousins}($s_p$, $I_i$, $\mathcal{DB}$) \Comment{Sample cousins; \cref{sec:cousin_sampling}}

        \State \textbf{// Multi-Agent-Aided Strategy Generation}
        \State $h \leftarrow$ \textsc{ResearchAgent}($s_p$, $\mathcal{C}$, $\mathcal{I}$) \Comment{Generate hypothesis}
        \State $(c, m) \leftarrow$ \textsc{CodingTeam}($h$, $s_p$, $\mathcal{C}$) \Comment{Generate code $c$ and backtest results $m$}
        \State $a \leftarrow$ \textsc{EvaluationTeam}($h$, $c$, $m$) \Comment{Analyze hypothesis, code, and results}

        \State \textbf{// Database Update}
        \State $s_{new} \leftarrow (h, c, m, a)$ \Comment{Compose complete strategy}
        \State $\mathbf{f} \leftarrow$ \textsc{ComputeFeatures}($s_{new}$) \Comment{Compute feature vector from backtest results}
        \State \textsc{UpdateDatabase}($s_{new}$, $\mathbf{f}$, $\mathcal{DB}$, $I_i$) \Comment{Update feature map and island}
        \State $\mathcal{I} \leftarrow \mathcal{I} \cup \{a\}$ \Comment{Accumulate insights for future generations}
    \EndFor

    \State \textbf{// Migration and Insight Management}
    \If{$g \bmod M = 0$ and $g > 0$}
        \State \textsc{MigrateStrategies}($\{I_1, \ldots, I_N\}$, $\mathcal{DB}$) \Comment{Migrate top 10\% from each island to neighbors}
    \EndIf
    \If{$g \bmod K = 0$ and $g > 0$}
        \State $\mathcal{I} \leftarrow$ \textsc{ManageInsights}($\mathcal{I}$) \Comment{Filter redundancy \& Curate and consolidate insights}
    \EndIf
\EndFor
\State \Return Evolutionary database $\mathcal{DB}$
\end{algorithmic}
\end{algorithm*}

\section{Feature Map of QuantEvolve}
\label{sec:feature_map}
The feature map organizes the strategy population as a multi-dimensional grid where each dimension represents a strategy attribute (e.g., Sharpe ratio, trading frequency, strategy category). Each dimension is discretized into bins, and their combination forms a \textit{feature vector} that uniquely identifies a cell. Each cell stores the best-performing strategy for that feature vector, ensuring that diverse behavioral profiles are preserved throughout evolution.

This structure provides two advantages: it enables personalized strategy recommendation by matching investor preferences to specific feature combinations, and it improves robustness by maintaining strategies optimized for different market regimes. The likelihood of discovering superior strategies also increases with greater population diversity.

\subsection{Feature Dimensions}

We select dimensions that align with common investor requirements as detailed in \cref{tab:feature_dims}.

\begin{table}[t]
\small
\centering
\begin{tabular}{@{}ll@{}}
\toprule
\textbf{Dimension} & \textbf{Description} \\
\midrule
Strategy Category & Momentum, mean-reversion, arbitrage, etc. \\
Trading Frequency & Number of trades per period \\
Maximum Drawdown & Largest peak-to-trough decline \\
Sharpe Ratio & Risk-adjusted return \\
Sortino Ratio & Downside risk-adjusted return \\
Total Return & Cumulative return \\
\bottomrule
\end{tabular}
\caption{Feature map dimensions}
\label{tab:feature_dims}
\end{table}

This design enables targeted matching: risk-averse investors seeking stable returns can select strategies with high Sharpe ratios and low maximum drawdown, while aggressive investors may prefer high trading frequency and total return.

The feature map is extensible to asset-specific requirements. For futures trading, dimensions such as rollover costs or roll yield can be added; for cryptocurrency markets, asset type categories (payment tokens, utility tokens, meme coins) can be incorporated.

\subsection{Feature Bins}
Each feature dimension is partitioned into discrete bins, with the number of bins per dimension determined by the desired granularity.
Continuous variables such as trading frequency or Sharpe ratio are discretized by equally dividing the range between the minimum and maximum values.

To maximize strategy diversity, we represent the strategy category dimension using binary encoding.

Given three strategy families—momentum, arbitrage, and mean-reversion—a strategy exhibiting both momentum and mean-reversion characteristics would be encoded as 101. This binary representation ensures that all possible category combinations occupy distinct bins, preventing strategies with different behavioral profiles from competing directly for the same cell .

\subsection{Feature Cells}
When a new strategy is generated, its characteristics are evaluated and mapped to a feature vector by determining the appropriate bin for each dimension.
This vector identifies a unique cell in the feature map.
The new strategy is added to that cell only if it outperforms the currently stored strategy; otherwise, the existing strategy is retained.
Although rejected strategies are not added to the feature map, we preserve them in a separate archive for potential use in future strategy generation. We refer to the feature map together with this archive as the \textit{Evolutionary Database}.

\subsection{Islands and Migration}

\begin{figure}[t]
    \centering
    \includegraphics[width=0.5\linewidth]{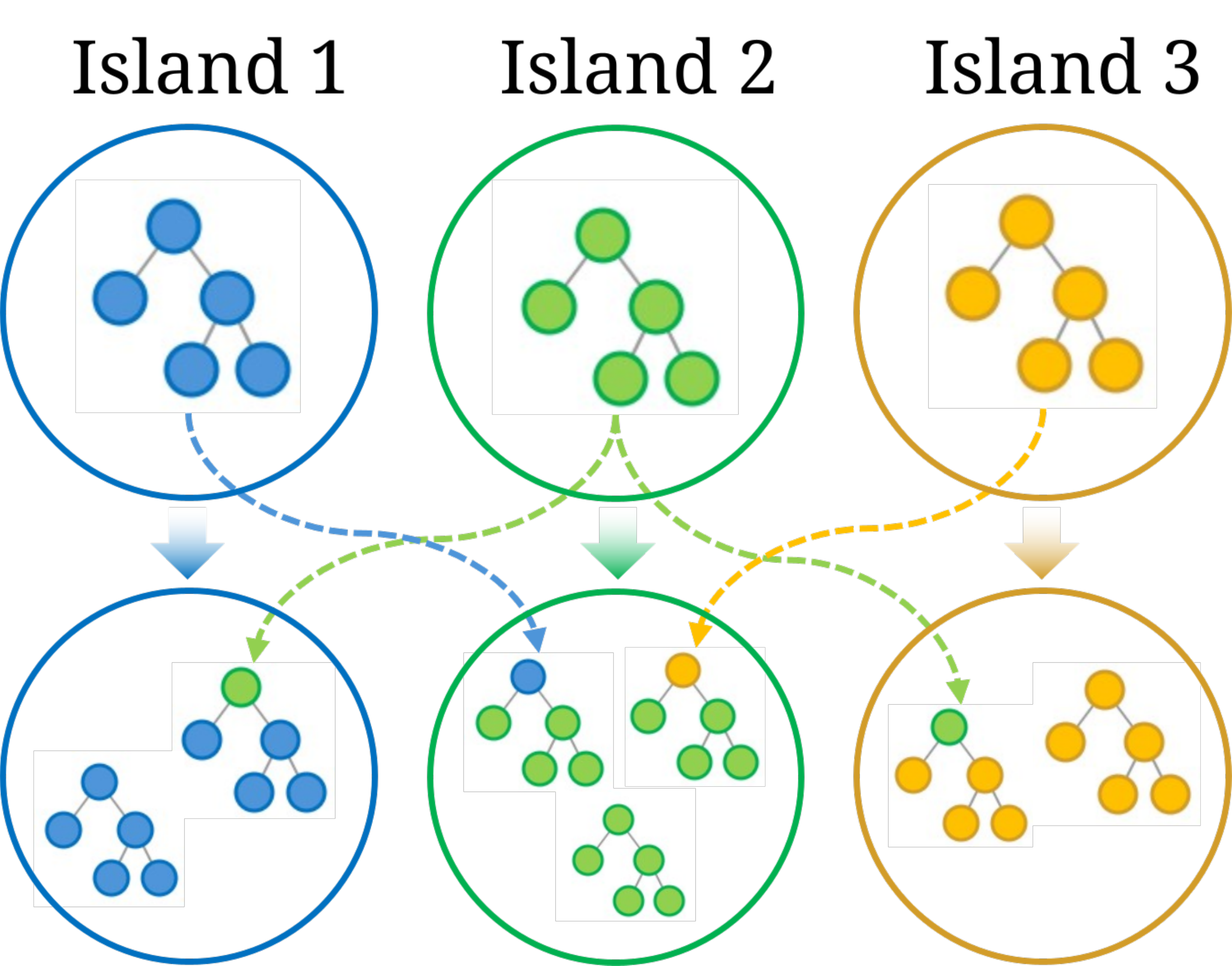}
    \Description{}
    \caption{Exchange of superior strategies across islands.}
    \label{fig:migration}
\end{figure}

To balance exploration and exploitation, we employ an island model where multiple populations evolve independently while sharing a common feature map. Each island is initialized with strategies from different categories (\cref{sec:initialization}) and evolves in isolation during early generations, developing specialized expertise in particular trading approaches.

Periodically, islands exchange their best-performing strategies through migration. This mechanism gradually enriches each population with diverse trading concepts, enabling the emergence of sophisticated hybrid strategies that combine characteristics from multiple approaches. Over time, the evolutionary focus shifts from depth within specialized categories to breadth across the entire strategy landscape (\cref{fig:migration}), producing increasingly complex and robust solutions.


\section{Multi-Agent System of QuantEvolve}
\label{sec:multi-agent}

To effectively explore the high-dimensional search space of trading strategies and discover high-performance solutions, we propose a novel multi-agent system that evolves strategies based on hypotheses, as illustrated in \cref{fig:multi-agent-arch}.
This multi-agent system generates new strategies following the process outlined below, using a parent strategy sampled from the feature map and cousin strategies with similar characteristics. 
We implement our multi-agent system using an ensemble of Qwen3-30B-A3B-Instruct-2507 (lightweight, faster responses) and Qwen3-Next-80B-A3B-Instruct (larger, more thoughtful analysis), balancing efficiency with reasoning depth~\cite{yang2025qwen3}.

\begin{enumerate}[leftmargin=1.5em]
    \item Hypothesis Generation: The research agent analyzes parent and cousin strategies to generate new hypotheses about trading strategies.
    \item Strategy Implementation: Based on the generated hypotheses, the coding team develops trading strategy code. They perform a backtest with the generated code and iteratively refine it if issues are identified.
    \item Evaluation and Analysis: The evaluation team analyzes the hypotheses, strategy code, and backtesting results to derive new insights.
    \item Strategy Storage: The generated strategy is stored in the appropriate cell of the feature map according to its characteristics. If a superior strategy already exists in that cell, the new strategy is rejected.
\end{enumerate}

\subsection{Initialization with Data Agent}
\label{sec:initialization}
\begin{figure}[t]
    \centering
    \includegraphics[width=0.8\linewidth]{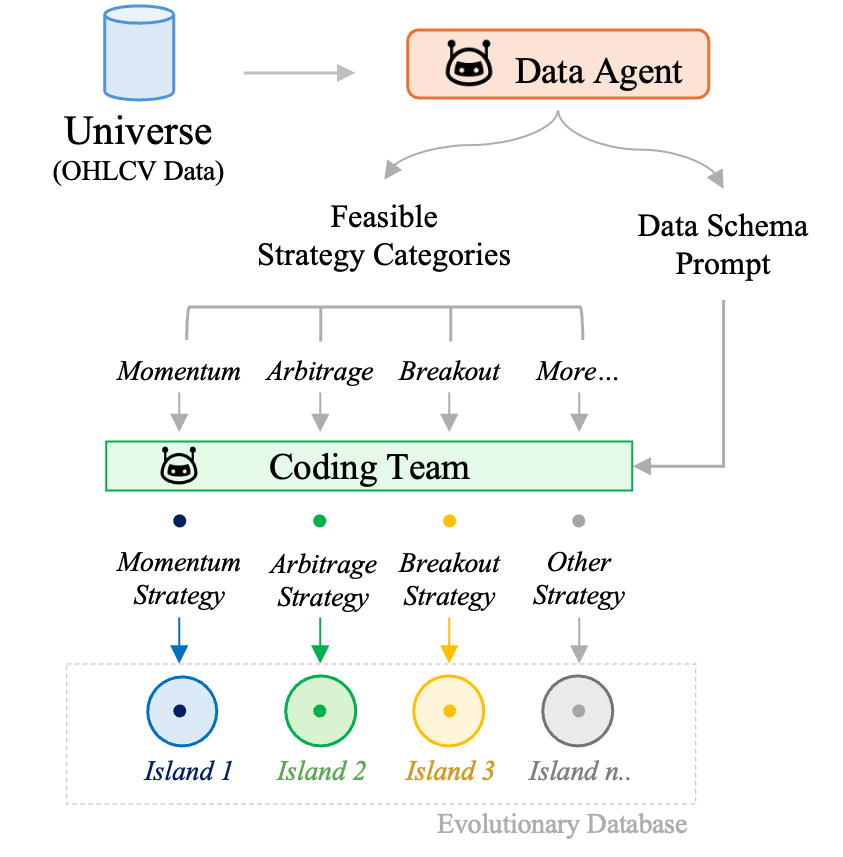}
    \caption{Initialization of QuantEvolve.}
    \label{fig:initial_setup}
\end{figure}

We initialize QuantEvolve by constructing an evolutionary database that serves as the foundation for subsequent optimization. 
As shown in \cref{fig:initial_setup}, we provide the \textit{Data Agent} with the available data universe—in our case, daily OHLCV data for six equities and two futures contracts.
The agent analyzes the input data structure, including file format (CSV, Parquet), schema, column definitions, and metadata required for reliable backtesting. Based on this analysis, the agent generates a \textit{Data Schema Prompt}, a structured specification that guides subsequent strategy design.

In parallel, the data agent identifies $C$ strategy categories derivable from the data universe, such as momentum, breakout, seasonality, and mean-reversion. For each category, the \textit{Coding Team} generates a simple but representative seed strategy using the schema prompt and category specification. For instance, the team produces a baseline momentum strategy for the momentum category and a prototypical breakout strategy for the breakout category. The team also generates a buy-and-hold benchmark, yielding $C+1$ initial strategies total.

Each of these $C+1$ strategies forms a separate island in the evolutionary database, representing distinct starting points for strategy evolution.
The initialization procedure thus constructs $N=C+1$ islands, with each island populated by a single seed strategy that serves as the foundation for subsequent evolutionary processes.
Subsequent evolutionary cycles iteratively reproduce, mutate, recombine, and migrate strategies within and across islands, enabling the emergence of increasingly sophisticated solutions.

\subsection{Parent and Cousins Sampling}


    


In our framework, reproduction is performed by sampling a parent strategy and cousin strategies that share similar characteristics with the parent.
The parent strategy serves as the primary reference for generating a new strategy, while the cousin strategies help to diversify its traits and explore novel combinations.

\subsubsection{Parent Strategy Selection}

When sampling a parent strategy, we balance exploitation of high-performing strategies with exploration of diverse characteristics to prevent premature convergence. We randomly select an island, then apply one of two sampling methods:
\begin{itemize}[leftmargin=1em]
    \item \textbf{Best parent:} Uniformly samples from strategies on the feature map within the selected island.
    \item \textbf{Diverse parent:} Uniformly samples from the entire population of the selected island.
\end{itemize}

Formally, let $I$ denote the selected island and $M_I \subseteq I$ represent strategies from island $I$ that exist on the feature map. The parent selection probability is:
\begin{equation}
P(s_p = s) = \begin{cases}
\frac{\alpha}{|M_I|} & \text{if } s \in M_I \text{ (best parent)} \\
\frac{1-\alpha}{|I|} & \text{if } s \in I \text{ (diverse parent)}
\end{cases}
\label{eq:parent_sampling}
\end{equation}

where $\alpha \in [0, 1]$ controls the exploitation-exploration trade-off. Higher $\alpha$ increases selection pressure toward high-performing strategies, while lower $\alpha$ promotes diversity. We use $\alpha = 0.5$ in our experiments to balance performance and diversity equally.

\subsubsection{Cousin Strategies Selection}
\label{sec:cousin_sampling}
To enrich the context for reproduction, we select not only a parent strategy but also other strategies with similar characteristics, which we refer to as cousin strategies.
To enhance strategy quality while promoting diversity, we select cousin strategies using three methods:
\begin{itemize}[leftmargin=1em]
    \item \textbf{Best Cousins:} High-performing strategies from the parent's island.
    \item \textbf{Diverse Cousins:} Strategies close to the parent in the feature space.
    \item \textbf{Random Cousins:} Strategies selected uniformly at random from the entire population of the parent's island.
\end{itemize}

We formalize diverse cousin selection as follows. 

Let $\mathbf{f}_p = (f_p^1, f_p^2, \ldots, f_p^D)$ denote the parent strategy's feature vector, where $D$ is the number of feature dimensions and $f_p^d$ represents the bin index for dimension $d$. For each dimension, we perturb the parent's feature bin to generate a neighbor feature vector $\mathbf{f}_c$:

\begin{equation}
f_c^d = \begin{cases}
\lfloor\mathcal{N}(f_p^d, \sigma_d^2)\rfloor & \text{if dimension } d \text{ is continuous} \\
\text{BitFlip}(f_p^d, k_{bf}) & \text{if dimension } d \text{ is strategy category}
\end{cases}
\end{equation}

where $\sigma_d$ controls the sampling radius for continuous dimension $d$, and $\text{BitFlip}(b, k_{bf})$ randomly selects and flips one bit from the binary encoding $b$, repeated $k_{bf}$ times. Note that the same bit can be selected multiple times, potentially canceling out previous flips. We then retrieve from the feature map the strategy with feature vector $\mathbf{f}_c$ as a diverse cousin.

For instance, given a parent with Sharpe ratio in bin 3 and strategy category bin '1001', we sample a diverse cousin by drawing the Sharpe ratio bin from $\lfloor\mathcal{N}(3, \sigma_d^2)\rfloor$ and performing $k_{bf}$ bitflips on '1001'. In our experiments, we set $\sigma_d = 1.0$ for all continuous dimensions and $k_{bf} = n/4$, where $n$ is the bit length of the strategy category. We select two best cousins, three diverse cousins, and two random cousins per parent.

\subsection{Research Agent}

The research agent generates new hypotheses for trading strategies based on insights from parent and cousin strategies. To formulate a hypothesis, we provide three key inputs to the agent: (1) the parent and cousin strategies' hypothesis, code, backtesting results, and analysis, (2) the Data Schema Prompt, and (3) insights accumulated from previous generations.
The agent constructs each hypothesis with the following components:
\begin{enumerate}[leftmargin=1em]
    \item \textbf{Hypothesis}: A testable statement grounded in financial theory, market dynamics, or statistical analysis that defines the core trading strategy.

    \item \textbf{Rationale}: Theoretical and empirical justification based on parent and cousin strategies, data patterns, or financial theories.

    \item \textbf{Objectives}: Specific quantitative and qualitative goals serving as evaluation criteria.

    \item \textbf{Expected Insights}: Anticipated learning outcomes and contributions to future strategy evolution.

    \item \textbf{Risks and Limitations}: Potential risks and limitations, including data bias, overfitting risks, and exceptional market conditions.

    \item \textbf{Experimentation Ideas}: Future directions for modifying, extending, or combining the hypothesis with other approaches.
\end{enumerate}

We adopt this hypothesis-driven method to address two fundamental challenges in evolutionary strategy generation. First, the trading strategy search space is exceptionally high-dimensional. Second, evolutionary frameworks naturally favor broad, shallow exploration over deep, focused refinement. Our approach provides directional guidance that enables systematic improvement while preserving diversity, allowing convergence toward superior performance within limited generations.

\subsection{Coding Team}
The coding team translates hypotheses into executable Python trading strategies. When a hypothesis presents well-defined logic, the team implements it directly; when further exploration is needed, the team develops experimental code to validate concepts before full implementation. The workflow follows four stages:

\begin{enumerate}[leftmargin=1.5em]
    \item \textbf{Initial Implementation}: We translate the hypothesis into Python code adhering to the data schema and backtesting framework, handling edge cases (missing data, execution constraints).
    
    \item \textbf{Backtesting}: We execute backtests to generate performance metrics (total return, Sharpe ratio, Sortino ratio, Maximum Drawdown, Trading frequency, Information Ratio).

    \item \textbf{Iterative Refinement}: When backtesting reveals issues—logical errors, performance anomalies, unexpected behaviors—we iteratively refine the code by debugging edge cases, optimizing efficiency, adjusting parameters, or adding risk constraints.
    
    \item \textbf{Performance Reporting}: Once the strategy passes backtesting without errors and demonstrates stable performance, we finalize the implementation and report code with metrics.
\end{enumerate}

We leverage Zipline~\cite{zipline_reloaded}, an open-source backtesting engine simulating market mechanics (slippage, commissions), and QuantStats~\cite{quantstats} for quantitative analysis and risk metrics.

\subsection{Evaluation Team}

The evaluation team analyzes hypotheses, code, and backtesting results to derive actionable insights guiding evolution. This team provides critical feedback, ensuring each generation's learnings are systematically captured. The analysis comprises five functions:

\begin{enumerate}[leftmargin=1.5em]
    \item \textbf{Hypothesis Analysis}: We evaluate hypothesis quality—assessing financial grounding, comprehensiveness (rationale, objectives, risks), testability, internal consistency, and market alignment.
    
    \item \textbf{Code Analysis}: We verify implementation fidelity to hypotheses, checking correct trading logic, proper edge case handling, code quality (\textit{e.g.}, readability, efficiency), and categorize strategies for feature map placement.

    \item \textbf{Backtest Analysis}: We treat results as hypothesis experiments, determining whether outcomes support, refute, or nuance the hypothesis. We identify successful/failed components, diagnose failure modes, and propose modifications or additional experiments (parameter optimization, alternative indicators, robustness tests).

    \item \textbf{Insight Extraction}: We synthesize actionable insights from hypotheses, code, and results—capturing success/failure patterns, promising directions, and unexplored strategy space areas. These insights directly inform future hypothesis generation.

    \item \textbf{Insight Management}: Every 50 generations, we curate accumulated insights—filtering redundancy, consolidating findings, and maintaining island-specific institutional memory to prevent repeated failures while reinforcing successful patterns.
\end{enumerate}

This systematic analysis creates a knowledge feedback loop accelerating evolution. By transforming results into structured insights, we enable the research agent to generate increasingly sophisticated hypotheses building on prior learnings. This cumulative knowledge acquisition distinguishes our approach from purely stochastic methods, enabling efficient convergence while maintaining population diversity.

%% file: sec/experimental_setup.tex
\section{Experimental Setup}

\subsection{Datasets}

\begin{table}[t]
    \centering
        \begin{tabular}{ll}
            \toprule
            \textbf{Market} & \textbf{Symbols} \\
            \midrule
            \textbf{Equities} & AAPL, NVDA, AMZN, GOOGL, MSFT, TSLA \\
            \textbf{Futures} & ES, NQ \\
            \bottomrule
    \end{tabular}
    \caption{Asset Universes.}
    \label{tab:asset_universe}
\end{table}

\begin{table}[t]
    \centering
    \begin{tabular}{llcc}
        \toprule
        \textbf{Market} & \textbf{Period} & \textbf{Date Range} & \textbf{Duration} \\
        \midrule
        \multirow{3}{*}{\textbf{Equities}} 
        & Training & 2015-08-01 to 2020-07-31 & 5 years \\
        & Validation & 2020-08-01 to 2022-07-31 & 2 years \\
        & Test & 2022-08-01 to 2025-07-31 & 3 years \\
        \midrule
        \multirow{3}{*}{\textbf{Futures}} 
        & Training & 2018-01-01 to 2021-07-31 & 3.6 years \\
        & Validation & 2021-08-01 to 2022-07-31 & 1 year \\
        & Test & 2022-08-01 to 2024-01-01 & 1.4 years \\
        \bottomrule
    \end{tabular}
    \caption{Temporal Data Split Configuration.}
    \label{tab:data_splits}
\end{table}

We evaluate QuantEvolve across two asset universes with different temporal configurations due to data availability constraints.
\Cref{tab:asset_universe} presents the asset universes, while \cref{tab:data_splits} details the temporal split configuration for each asset universe.
To ensure realistic backtesting conditions, we simulate per-share commission costs of \$0.0075 with a minimum of \$1.00 per trade, and implement volume-based slippage as a quadratic function of percentage of historical volume.

All results are reported using a train-validation-test framework: strategies evolve on training data, the best-performing strategy on the validation set (measured by combined score) is selected, and final performance is evaluated on the held-out test period.

\subsection{Baseline Strategies}
For equities, we construct a market capitalization-weighted portfolio of the six stocks rebalanced monthly, and an equal-weighted portfolio strategy rebalanced daily. We also compare against three technical indicator-based strategies: Risk Parity (RP), Relative Strength Index (RSI), and Moving Average Convergence Divergence (MACD). These strategies represent common quantitative approaches and serve as active baselines. All strategies, including baselines and evolved strategies, apply identical transaction costs (0.075\% commission) to ensure comparability.


For futures, we use individual buy-and-hold baselines for ES and NQ contracts. Each contract is sized with equal notional allocation at initialization based on price and point value, then held throughout the evaluation period. We report the average performance across both contracts as the aggregate baseline. While unified futures portfolio benchmarks are impractical due to heterogeneous contract specifications, this approach provides a fair reference for evaluating per-contract performance.

\subsection{Evaluation Metrics}
We evaluate strategies using four standard quantitative metrics and a composite objective for evolutionary selection.

\subsubsection{Individual Metrics.}
\begin{itemize}[leftmargin=1em]
    \item \textbf{Sharpe Ratio (SR)}: Risk-adjusted return measuring excess return per unit of volatility: $SR = \frac{\bar{R_p} - R_f}{\sigma_p}$, where $\bar{R_p}$ is average portfolio return, $R_f$ is the risk-free rate (set to 0), and $\sigma_p$ is return standard deviation.
    
    \item \textbf{Sortino Ratio (SOR)}: Similar to Sharpe but penalizes only downside volatility: $SOR = \frac{\bar{R_p} - R_f}{\sigma_d}$, where $\sigma_d$ is the standard deviation of negative returns.
    
    \item \textbf{Information Ratio (IR)}: Benchmark-relative performance: $IR = \frac{\bar{R_p} - \bar{R_b}}{\sigma_{p-b}}$, where $\bar{R_b}$ is benchmark return and $\sigma_{p-b}$ is tracking error.
    
    \item \textbf{Maximum Drawdown (MDD)}: Largest peak-to-trough decline: $MDD = \min_{t} \left(\frac{\text{Trough}_t - \text{Peak}_t}{\text{Peak}_t}\right)$.
\end{itemize}

\subsubsection{Combined Score for Evolution.}
To guide evolutionary selection, we construct a composite objective balancing profitability, benchmark outperformance, and downside risk:
\begin{equation}
    \text{Score} = SR + IR + MDD
    \label{eq:combined_score}
\end{equation}

This formulation rewards both absolute (SR) and relative (IR) risk-adjusted returns while penalizing drawdown severity. The equal weighting (1:1:1) avoids overemphasizing any single dimension, encouraging strategies that balance consistent alpha generation with capital preservation. This composite objective transforms evolution from pure return maximization into a robust measure of long-term viability across market regimes.

%% file: sec/result.tex
\section{Results in Equity Markets}

\subsection{Evolution of the Feature Map}
\begin{figure*}[t]
    \centering
    \captionsetup[subfigure]{font=scriptsize}
    \begin{subfigure}[b]{0.24\linewidth}
        \centering
        \includegraphics[trim={0.5cm 0.5cm 0.5cm 0.5cm}, clip, width=1.0\linewidth]{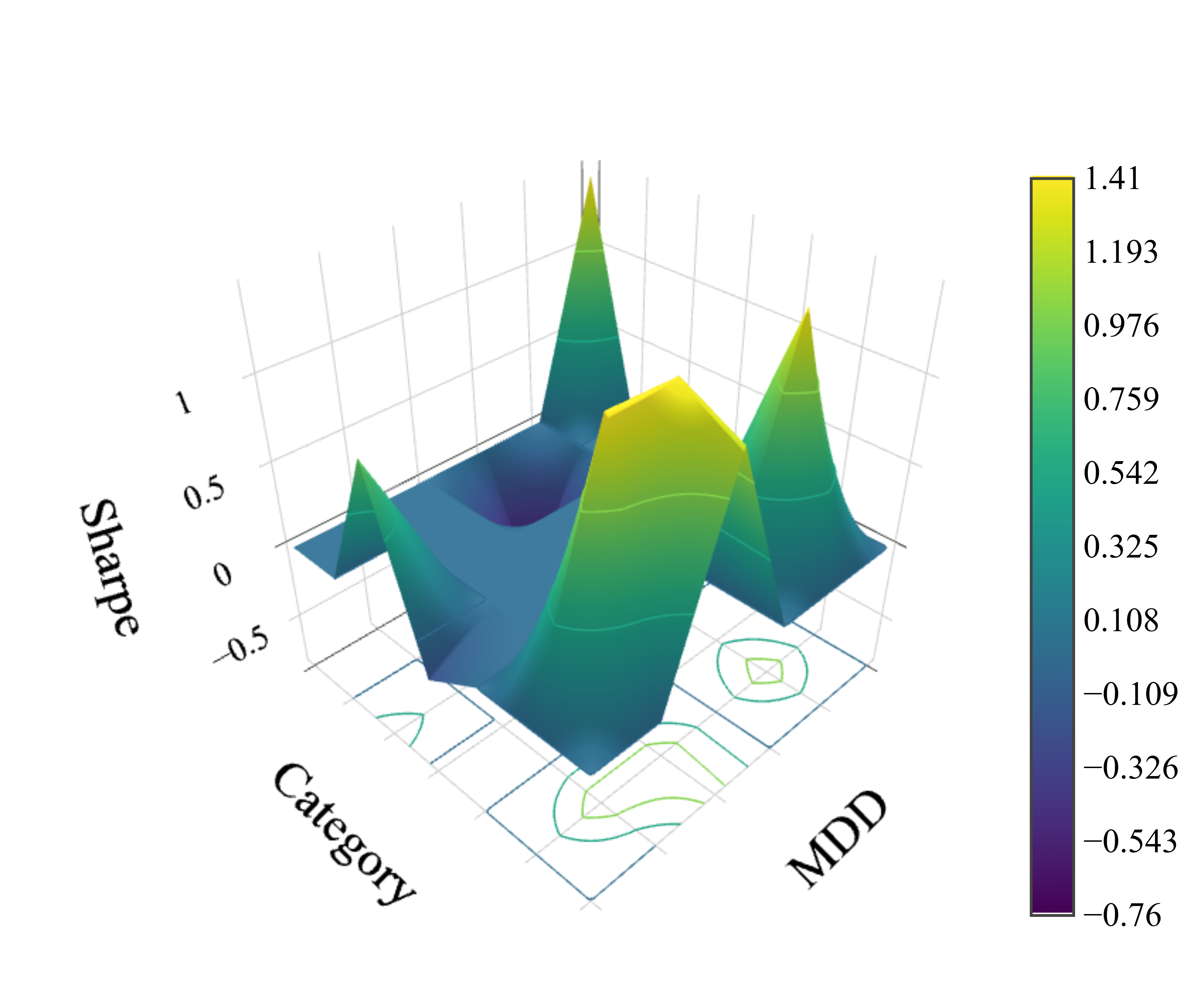}
        \subcaption{Gen 0}
    \end{subfigure}
    \hfill
    \begin{subfigure}[b]{0.24\linewidth}
        \centering
        \includegraphics[trim={0.5cm 0.5cm 0.5cm 0.5cm}, clip, width=1.0\linewidth]{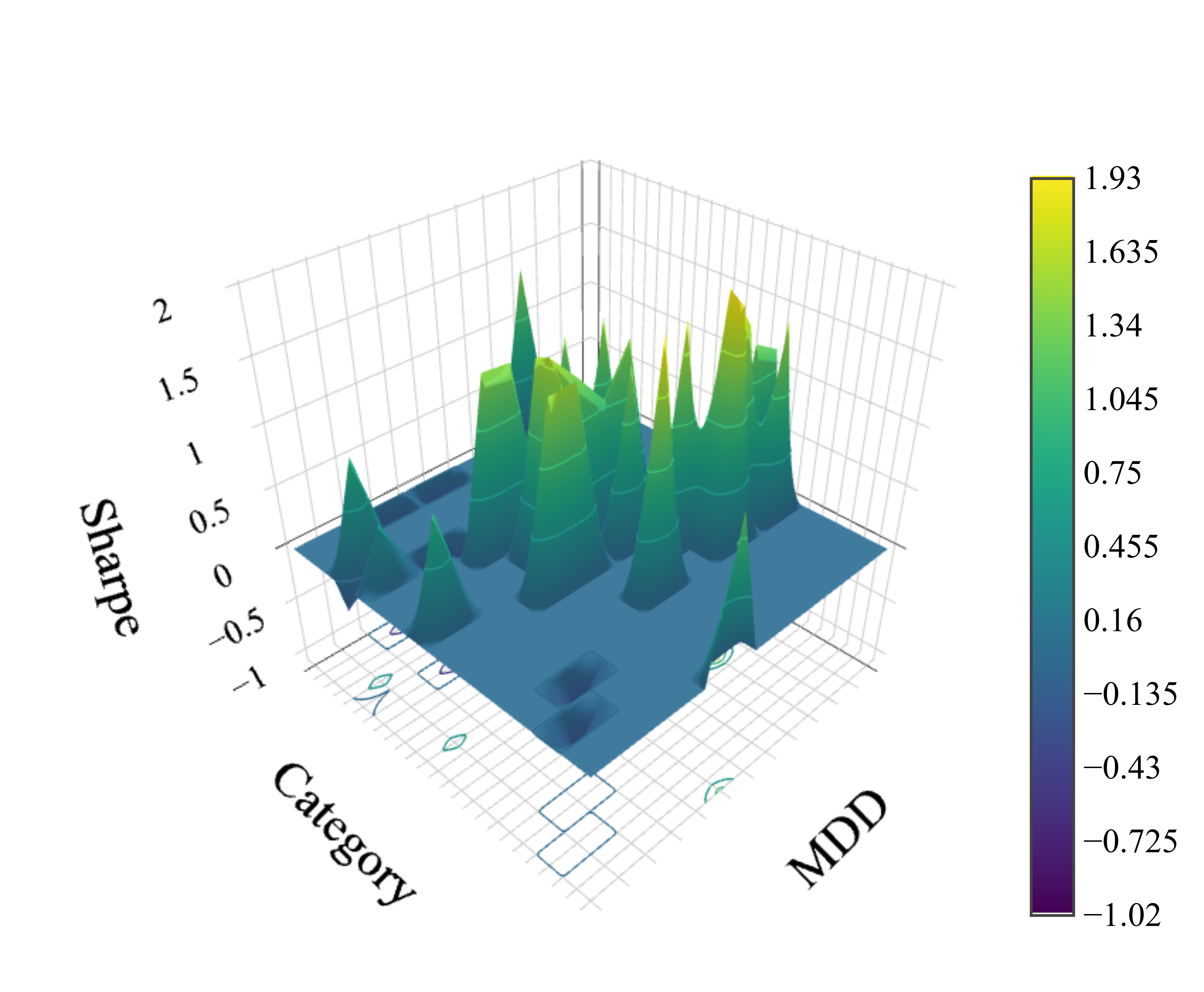}
        \subcaption{Gen 50}
    \end{subfigure}
    \hfill
    \begin{subfigure}[b]{0.24\linewidth}
        \centering
        \includegraphics[trim={0.5cm 0.5cm 0.5cm 0.5cm}, clip, width=1.0\linewidth]{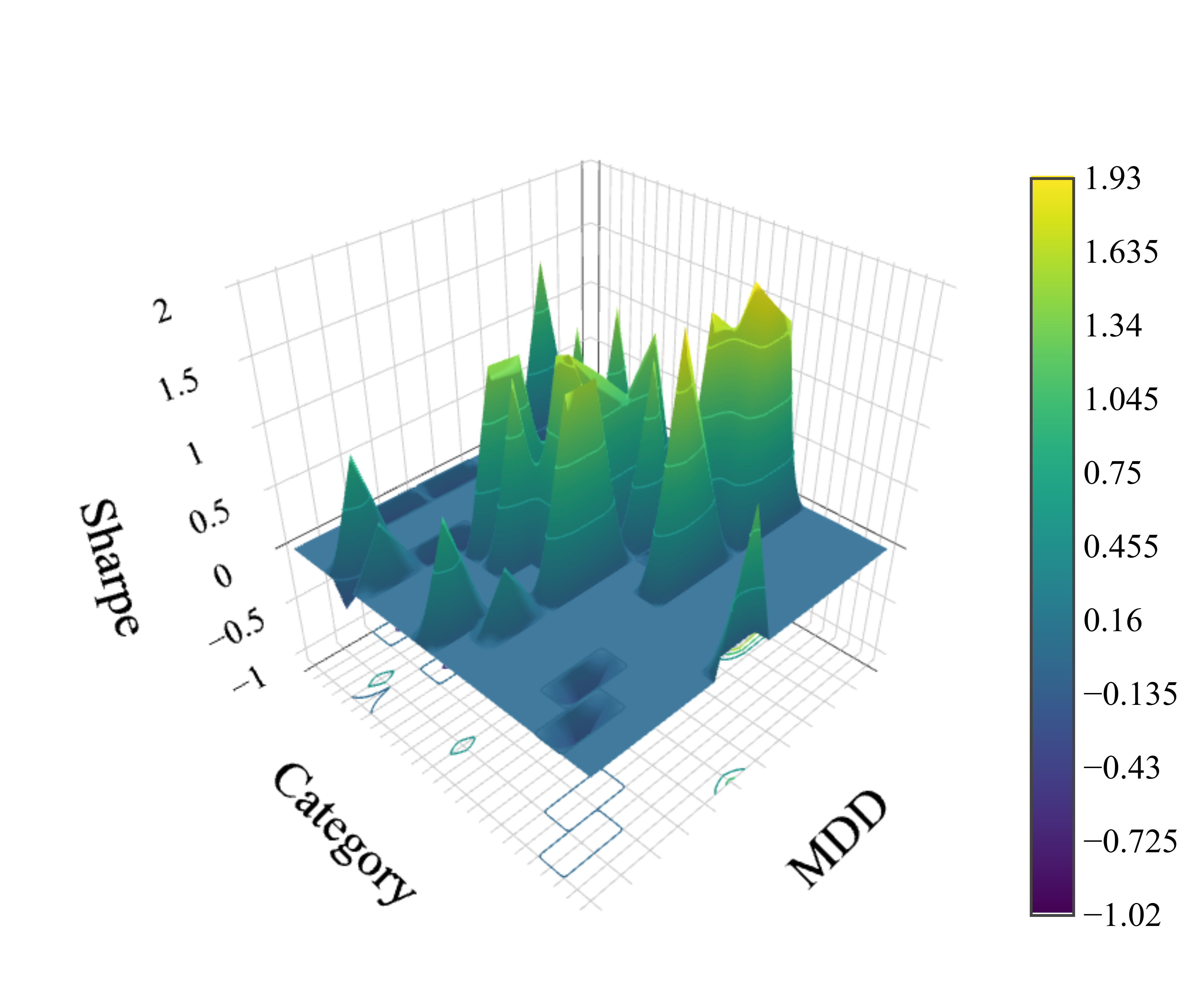}
        \subcaption{Gen 100}
    \end{subfigure}
    \hfill
    \begin{subfigure}[b]{0.24\linewidth}
        \centering
        \includegraphics[trim={0.5cm 0.5cm 0.5cm 0.5cm}, clip, width=1.0\linewidth]{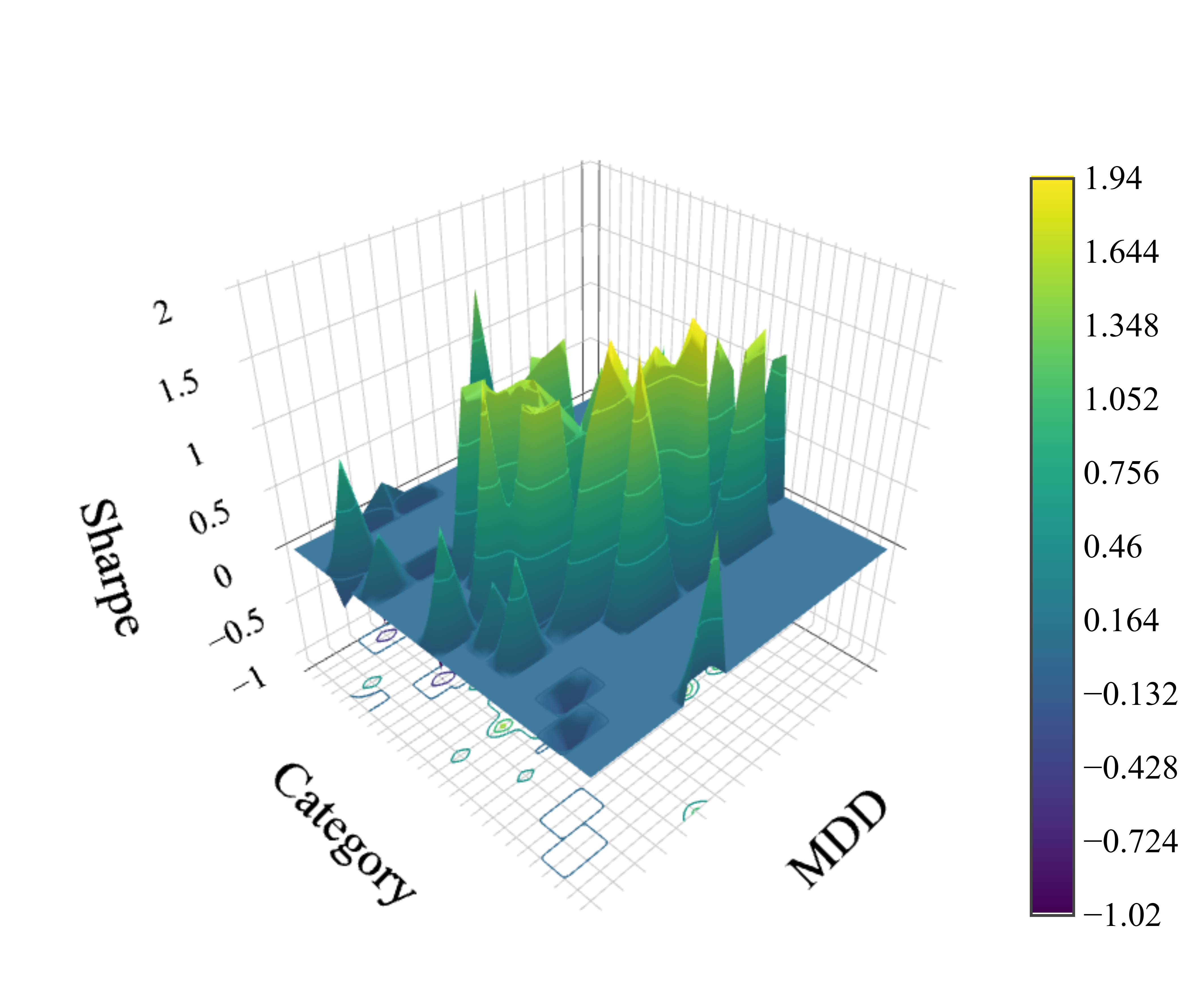}
        \subcaption{Gen 150}
    \end{subfigure}
    \vspace{-0.5em}
    \caption{
        Evolution of the feature map across max drawdown and category dimensions, measured by Sharpe ratio.
    }
    \label{fig:progress_featuremap_main}
\end{figure*}

\Cref{fig:progress_featuremap_main} presents three-dimensional projections of the feature map, enabling us to track its evolution across generations.
We visualize the feature map using strategy category and max drawdown as the two axes, where each cell displays the strategy with the highest Sharpe ratio among all strategies sharing the same category and max drawdown bin.

At generation 0, the feature map consists exclusively of a buy-and-hold baseline and representative strategies from each predefined category, the details of which are provided in the \cref{tab:strategy_category_equity}.
As evolution progresses, the cells of the feature map gradually become populated with increasingly diverse strategies, and the performance within each cell shows consistent improvement. This visualization demonstrates both the exploratory capacity of the evolutionary process—filling previously empty regions of the strategy space—and its ability to refine solution quality within each niche over successive generations.

\begin{figure*}[t]
    \centering
    \captionsetup[subfigure]{font=scriptsize}
    \begin{subfigure}[b]{0.24\linewidth}
        \centering
        \includegraphics[trim={0.5cm 0.5cm 0.5cm 0.5cm}, clip, width=1.0\linewidth]{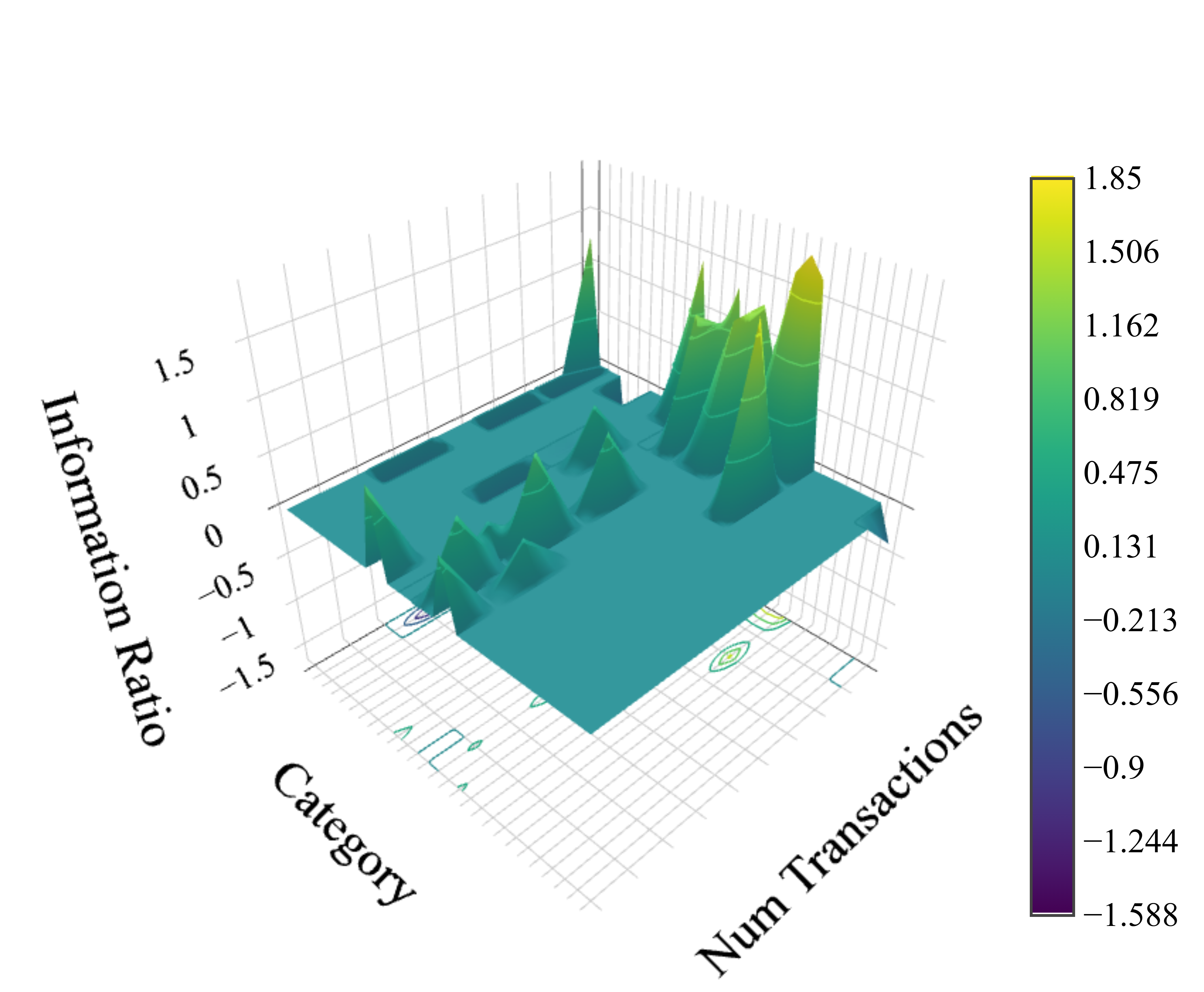}
        \subcaption{Num Transactions \& Category}
    \end{subfigure}
    \hfill
    \begin{subfigure}[b]{0.24\linewidth}
        \centering
        \includegraphics[trim={0.5cm 0.5cm 0.5cm 0.5cm}, clip, width=1.0\linewidth]{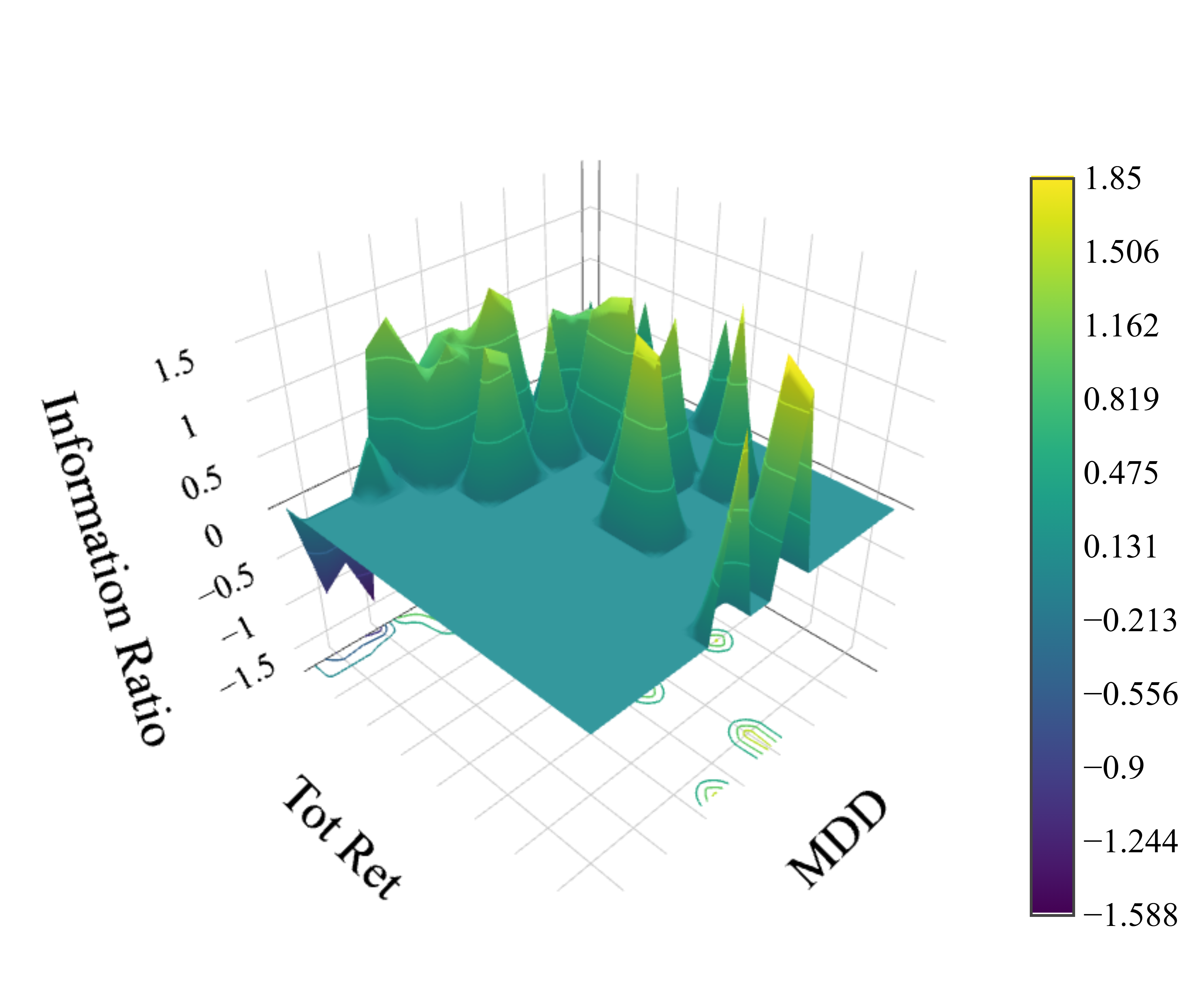}
        \subcaption{MDD \& Returns}
    \end{subfigure}
    \hfill
    \begin{subfigure}[b]{0.24\linewidth}
        \centering
        \includegraphics[trim={0.5cm 0.5cm 0.5cm 0.5cm}, clip, width=1.0\linewidth]{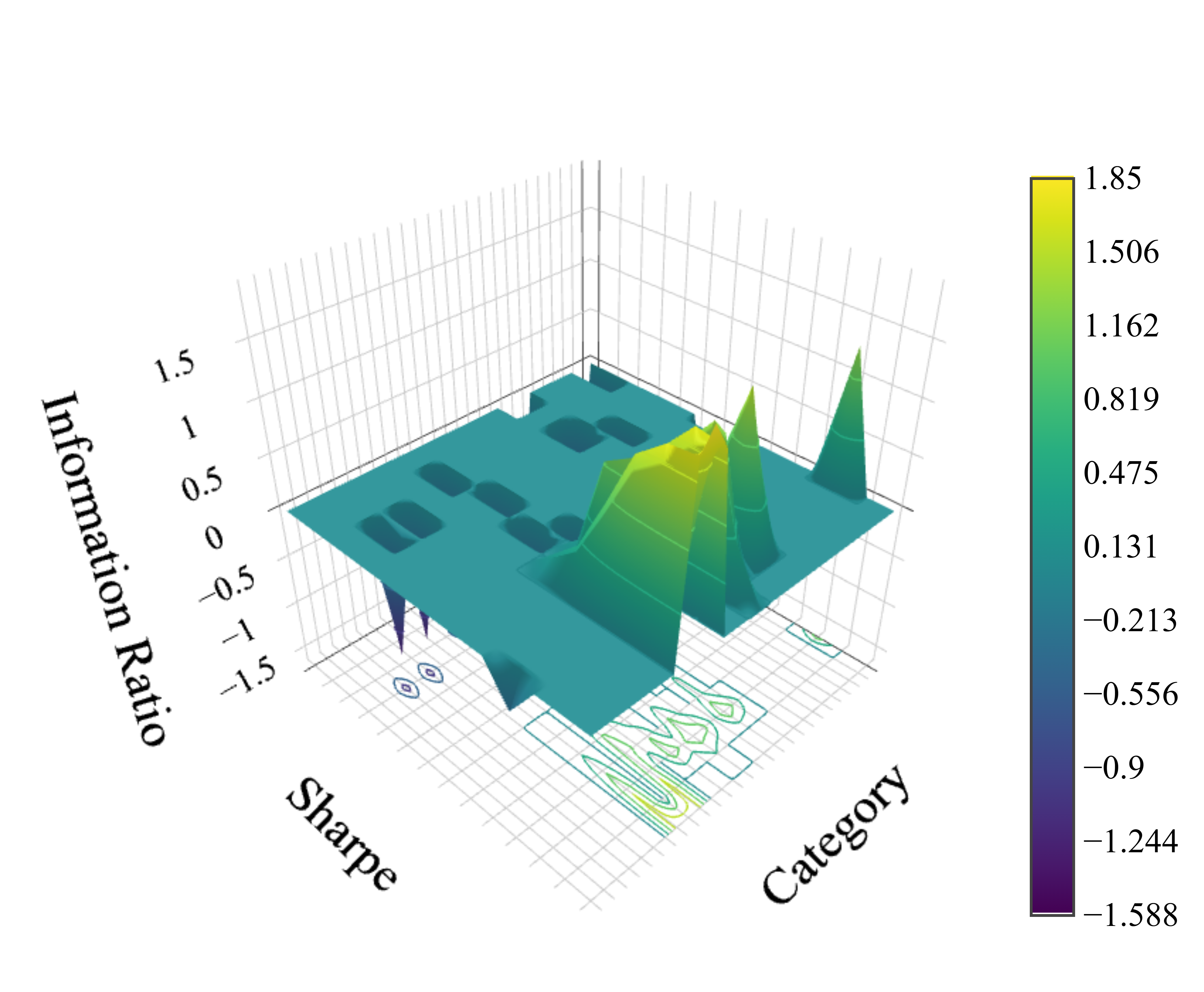}
        \subcaption{Category \& SR}
    \end{subfigure}
    \hfill
    \begin{subfigure}[b]{0.24\linewidth}
        \centering
        \includegraphics[trim={0.5cm 0.5cm 0.5cm 0.5cm}, clip, width=1.0\linewidth]{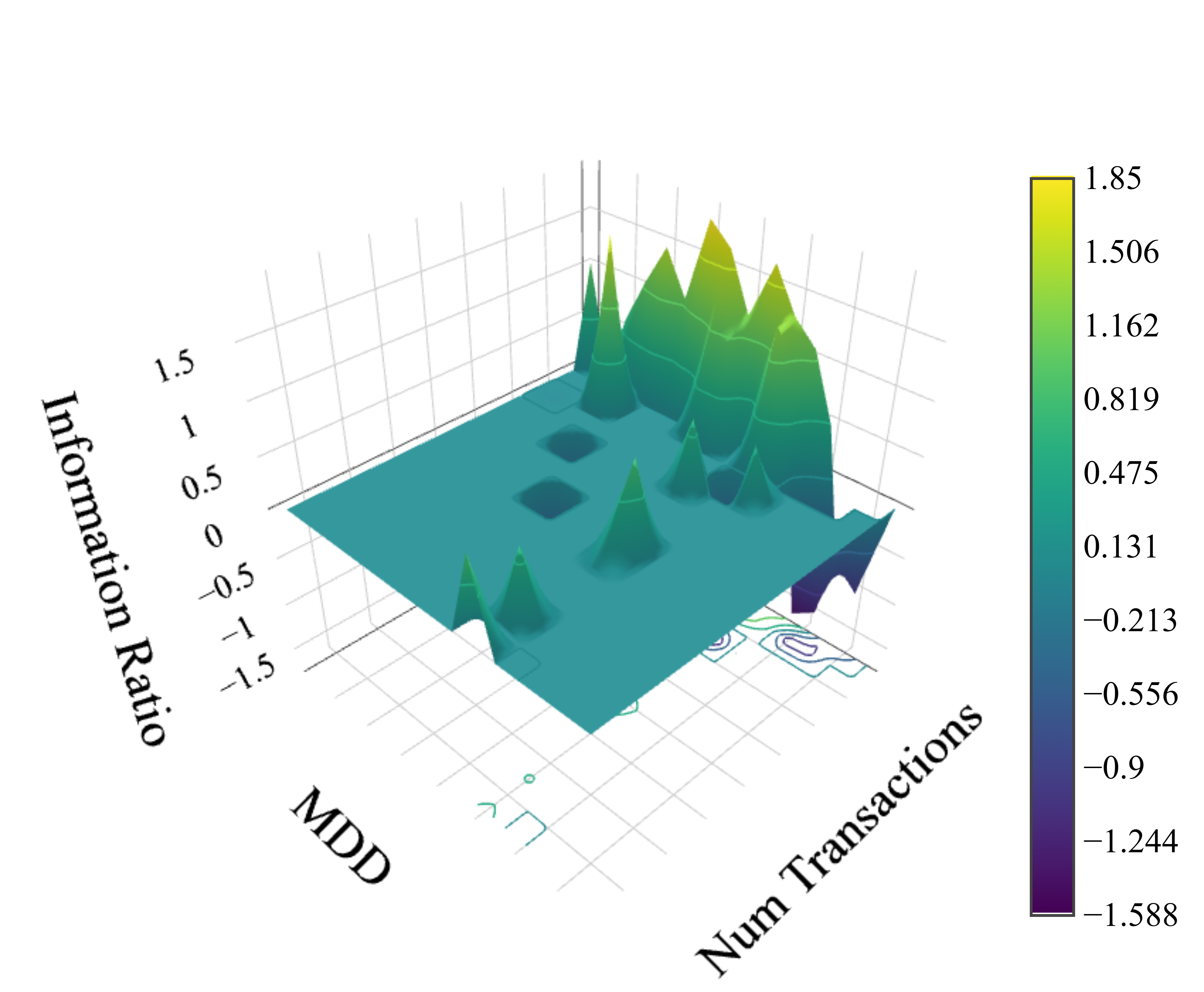}
        \subcaption{MDD \& Num Transactions}
    \end{subfigure}
    \vspace{-0.5em}
    \caption{
        Feature map visualizations across different dimension pairs, including the number of transactions and cumulative return, evaluated using the information ratio.
    }
    \label{fig:final_featuremap_main}
\end{figure*}

\Cref{fig:final_featuremap_main} presents the final generation's feature map visualizations across different dimension pairs, including number of transactions and cumulative return, among others, evaluated using the information ratio.
Several noteworthy patterns emerge from these visualizations.
First, strategies with higher cumulative returns tend to exhibit moderate levels of maximum drawdown rather than extreme values at either end of the spectrum.
This inverted U-shaped relationship suggests a trade-off between risk-taking and return generation: strategies with excessively low maximum drawdown appear to operate conservatively, resulting in limited upside potential, while those with very high maximum drawdown may expose portfolios to excessive downside risk that undermines long-term compounding.
Second, we observe a positive association between Sharpe ratio and information ratio across the feature map, which is intuitive given that both metrics evaluate risk-adjusted performance, albeit with different benchmark specifications.

These patterns indicate that the evolutionary process successfully identifies a diverse set of strategies occupying distinct regions of the risk-return space, rather than converging prematurely to a narrow subset of solutions.

\subsection{Evolution of the Strategy}

\begin{figure}[t]
    \centering
    \includegraphics[width=0.95\columnwidth]{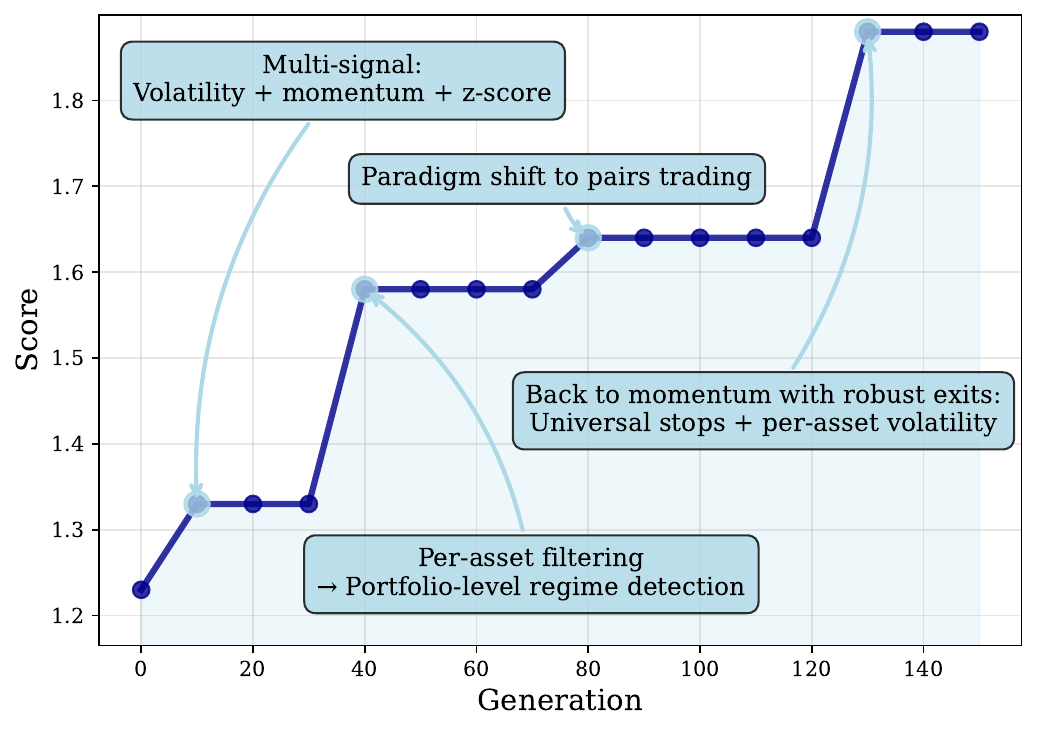}
    \caption{
        The evolutionary process leading to the optimal strategy in equity markets.
    }
    \label{fig:progress_strategy_main}
\end{figure}

\Cref{fig:progress_strategy_main} illustrates the evolutionary trajectory from generation 0 to 130, showing how the framework explores different strategy designs through iterative hypothesis testing.
The initial strategy (generation 0) employed volume-momentum signals.
Generation 10 expanded the signal space by incorporating momentum analysis across multiple timeframes, volatility filtering, and mean-reversion components.
While this approach increased cumulative returns, the maximum drawdown expanded.
Analysis suggested that the primary limitation had shifted from signal diversity to portfolio-level risk management: uniform weighting across assets did not account for differing volatility characteristics, and periodic rebalancing generated trades regardless of market conditions.

Generation 40 redirected focus toward risk management design.
Rather than continuing to refine individual signals, this iteration introduced portfolio-wide volatility monitoring.
It separated exit decisions from the monthly rebalancing schedule, enabling more responsive position management during adverse market movements.
These changes reduced the maximum drawdown compared to generation 10, although the improvement in the Sharpe ratio remained modest.

Generation 80 explored an alternative approach based on cointegration pairs trading, testing whether relative-value strategies could offer different risk-return characteristics.
This exploration failed to generate positive returns, primarily due to the unstable cointegration relationships during volatile periods.

Generation 130 selectively integrated insights from previous generations while maintaining implementation simplicity.
The strategy retained volume-momentum signals as the core component, incorporated per-asset volatility adjustments to address the uniform weighting limitation identified in generation 10, and adopted continuous position monitoring rather than schedule-dependent exits as explored in generation 40.
Notably, the framework avoided the complexity that contributed to the failure of generation 80.
The resulting strategy is comprised of three straightforward components: momentum-based entry signals, volatility-scaled position sizing, and trailing stop-loss rules.
This design contrasts with accumulating all features from each iteration, which would increase the parameter count and risk of overfitting.

This trajectory illustrates that strategy evolution in our framework proceeds through the exploration of different design choices, the selective retention of practical components, and the prioritization of robustness over architectural complexity.

\subsection{Evolution of the Insights}

\begin{figure}[t]
    \centering
    \includegraphics[width=0.95\columnwidth]{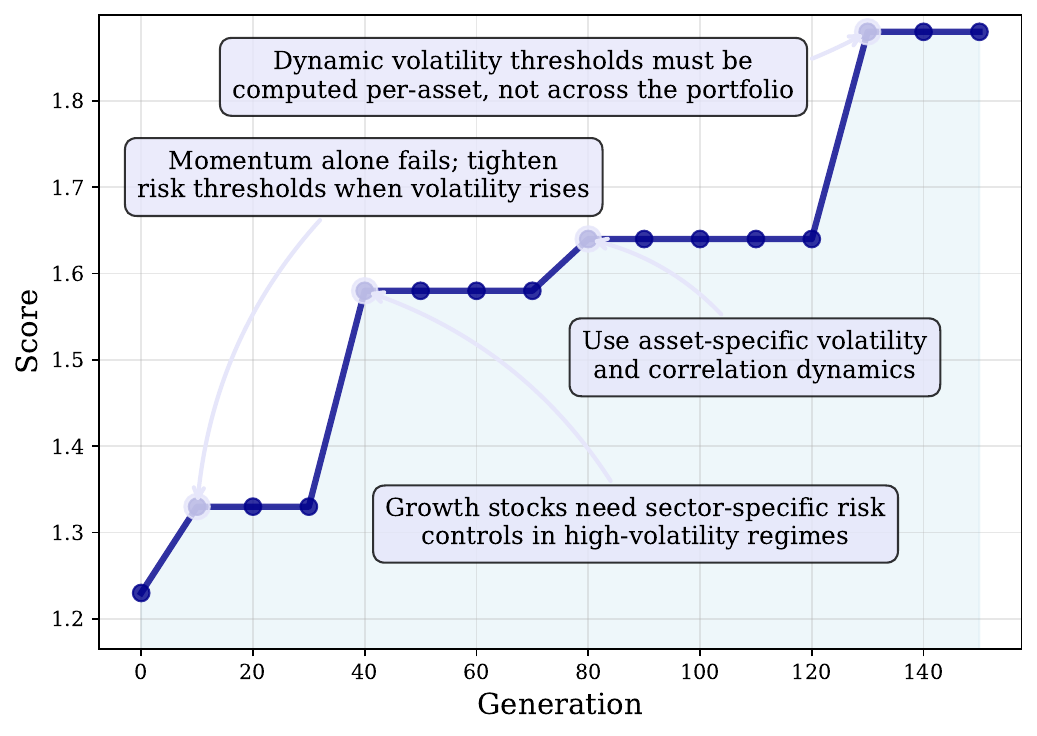}
    \caption{
        The evolutionary process of the insights in equity markets.
    }
    \label{fig:progress_insight_main}
\end{figure}

\Cref{fig:progress_insight_main} presents the evolutionary process of the insights in equity markets.
Generation 10 identified that simple volatility indicators fail during market crises, and recognized the need to combine volatility z-scores with VIX or moving averages.
Additionally, an alternative hypothesis emerged proposing sector diversification instead of signal refinement.
This diversity of approaches prevented premature convergence on a single solution.

Generations 40-80 yielded specialized insights.
For example, building on generation 10's observations, generation 40 found that growth stocks require sector-specific risk controls in high-volatility regimes.
As migration progressed, the framework began to combine previously separate findings: VIX integration, asset-specific thresholds, and volatility modeling emerged together in unified insights.
It also began distinguishing implementation errors from conceptual failures.
For instance, it determined that regime detection mechanisms failed due to incorrect volatility annualization rather than flawed logic.
Generation 80 integrated per-asset volatility thresholds with correlation effects, while new directions emerged investigating momentum persistence and market microstructure.

Generation 130 demonstrates insight consolidation and refinement.
The framework documented over 30 failed approaches, including returns-based volatility detection, sector moving average filters, and various covariance estimation methods.
Generation 10's insights evolved into detailed implementation specifications, such as requiring dynamic volatility thresholds to be computed per-asset rather than across the portfolio.
Through experimental validation, the framework found several techniques used in quantitative finance practice, including multi-signal confirmation, Kalman filtering for beta estimation, and risk parity scaling.

\subsection{Performance of the Strategies}

\begin{table}[t]
    \centering
    \small
        \begin{tabular}{ll|cccc}
            \toprule
            
            \multicolumn{2}{c}{\textbf{Model}} 
            & \multicolumn{1}{c}{SR} & \multicolumn{1}{c}{MDD} & \multicolumn{1}{c}{IR} & \multicolumn{1}{c}{CR}
            \\
            
            \midrule[0.25pt]
            
            \textit{Baselines}
            & MarketCap
            & 0.99 & -33\% & - & 99\%
            \\
            & Equal
            & 1.07 & -36\% & 0.80 & 129\%
            \\
            & MACD 
            & 1.10 & -39\% & 0.75 & 171\%
            \\
            & RSI
            & 1.03 & -37\% & 0.39 & 136\%
            \\
            & RP 
            & 1.22 & \textbf{-29\%} & 0.44 & 130\%
            \\
            
            \midrule[0.25pt]
            
            \textit{\textbf{Ours}}
            & Gen   0
            & 0.99 & -35\% & 0.03 & 100\%
            \\
            & Gen  50
            & 1.07 & -34\% & 0.49 & 119\%
            \\
            & Gen 100
            & 1.11 & -32\% & \textbf{0.87} & 128\%
            \\
            & \textbf{Gen 150}
            & \textbf{1.52} & -32\% & 0.69 & \textbf{256\%}
            \\
            
            \bottomrule
        \end{tabular}
    \caption{
        Experimental results in equity markets. MarketCap represents a portfolio strategy that rebalances monthly with each stock weighted by its market capitalization. We use this strategy as the benchmark for calculating the information ratio. Equal represents an equal-weighted portfolio strategy that rebalances daily. For detailed implementation of each baseline, please refer to the \cref{sec:baseline_code}.
    }
    \label{tab:performance_main}
\end{table}

In our study, we benchmarked the proposed framework against a set of widely adopted portfolio management approaches.
The baseline methods comprised two fundamental weighting schemes: MarketCap, a market capitalization-weighted portfolio rebalanced monthly that serves as the benchmark for information ratio calculations, and Equal, an equal-weighted portfolio rebalanced daily.
Additionally, we included the Risk Parity strategy (RP), which equalizes risk contributions across assets through inverse-volatility weighting, and two indicator-driven strategies: the Relative Strength Index (RSI), a contrarian mean-reversion framework, and the Moving Average Convergence Divergence (MACD), a momentum-following rule based on moving average crossovers.
These baselines represent complementary investment paradigms—passive indexing, risk balancing, momentum persistence, and mean reversion—providing a rigorous basis for comparison with the strategies generated by our QuantEvolve framework.

\Cref{tab:performance_main} shows that our framework achieves competitive out-of-sample test performance, demonstrating favorable results across both SR and CR relative to the baseline strategies.
These results suggest that evolutionary frameworks can be effectively applied to quantitative research and automated strategy generation.


\section{Results in Futures Markets}

\subsection{Evolution of the Strategy}

\begin{figure}[!t]
    \centering
    \includegraphics[width=0.95\columnwidth]{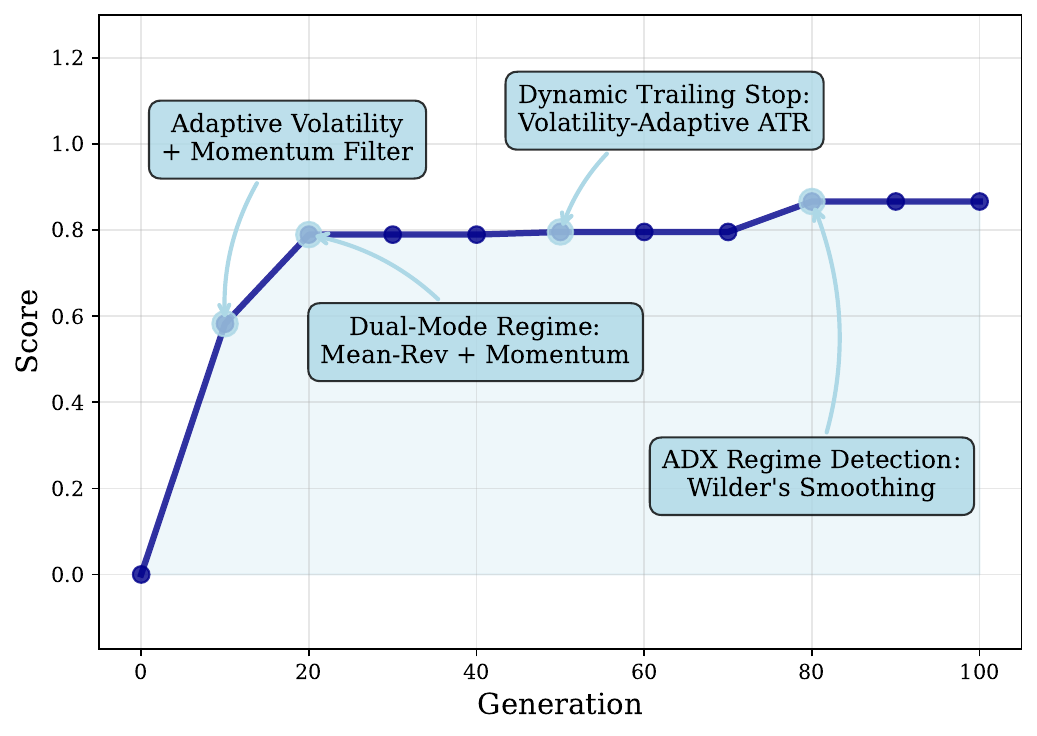}
    \caption{
        Strategy evolution in futures markets from generation 0 to 100.
    }
    \label{fig:strategy_evol_futures}
\end{figure}

\Cref{fig:strategy_evol_futures} illustrates the evolutionary trajectory in ES futures trading.
Generation 0 employed static mean reversion with fixed Bollinger Band thresholds and time-based exits, suffering from the inability to adapt to regime changes.

Generation 10 introduced dynamic volatility scaling and momentum confirmation. By calculating adaptive Z-scores from rolling volatility and requiring directional momentum alignment, it prevented counter-trend trades during strong moves. ATR-based position sizing and contract roll management substantially reduced drawdown, though performance remained inconsistent across regimes.

Generation 20 implemented a dual-mode system with explicit regime detection. During low-volatility periods, it applied mean-reversion with adaptive Bollinger Bands (2$\sigma$ to 3$\sigma$) and volume confirmation; during high-volatility regimes, it switched to momentum-following with trend filters. Rolling ATR quantiles and sigmoid transitions enabled smooth regime adaptation.

Despite evolution continuing through generation 100, generation 20 achieved superior out-of-sample performance. Later generations introduced increasingly complex mechanisms—ADX regime detection, multi-signal systems—that degraded generalization.
Generation 20's balance between adaptive sophistication and implementation parsimony exemplifies successful navigation of the bias-variance tradeoff.

\subsection{Evolution of the Insights}

\begin{figure}[!t]
    \centering
    \includegraphics[width=0.95\columnwidth]{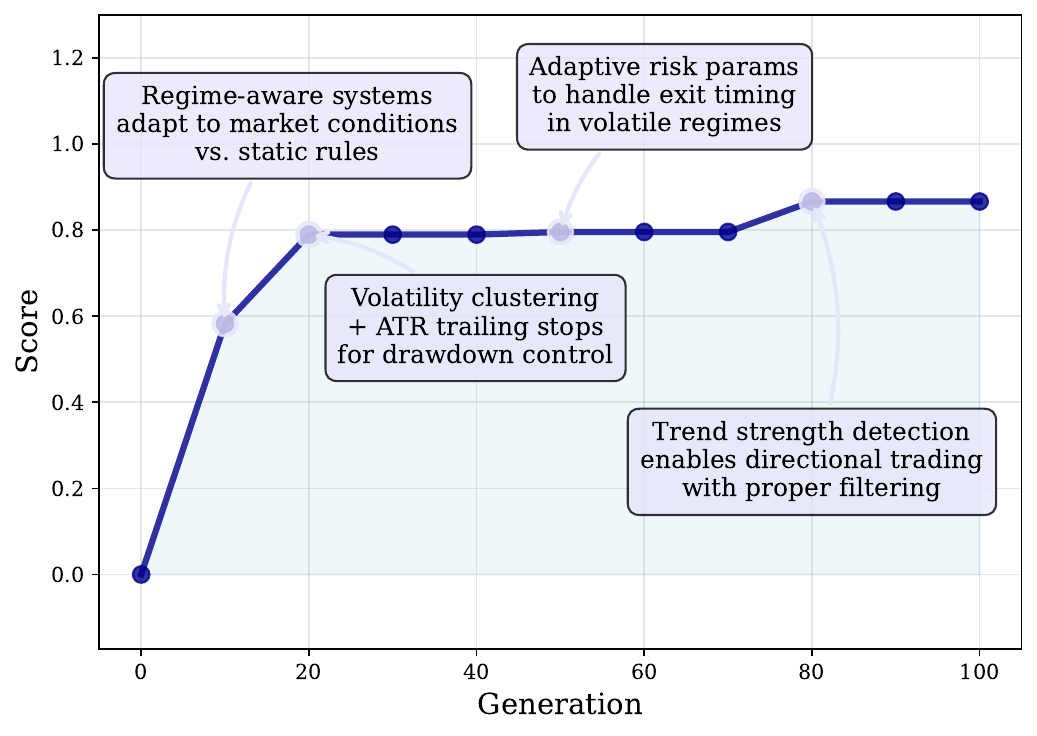}
    \caption{
        Insight evolution in futures markets.
    }
    \label{fig:insights_evol_futures}
\end{figure}

\Cref{fig:insights_evol_futures} shows the accumulation of strategic knowledge. Generation 10 identified that static volatility thresholds fail during regime transitions and documented that ATR-based position sizing reduces drawdowns more effectively than fixed contract allocation. It also recognized the importance of momentum confirmation to avoid premature mean-reversion entries.

Generation 20 synthesized these findings into regime-specific insights: low-volatility environments favor mean-reversion with tight bands, while high-volatility periods require momentum-following with wider thresholds. It discovered that volume confirmation prevents false signals during low-liquidity periods and that trailing stops should scale with current ATR rather than entry-time volatility. Critically, it documented failed approaches including returns-based regime detection and fixed holding periods, preventing future iterations from repeating these errors.

By generation 20, the framework had established that regime-aware systems outperform static rules in futures markets, that smooth transitions between modes reduce whipsaws, and that implementation simplicity often trumps theoretical sophistication.

\subsection{Performance on Futures Markets}

\begin{figure}[t]
    \centering
    \includegraphics[width=\linewidth]{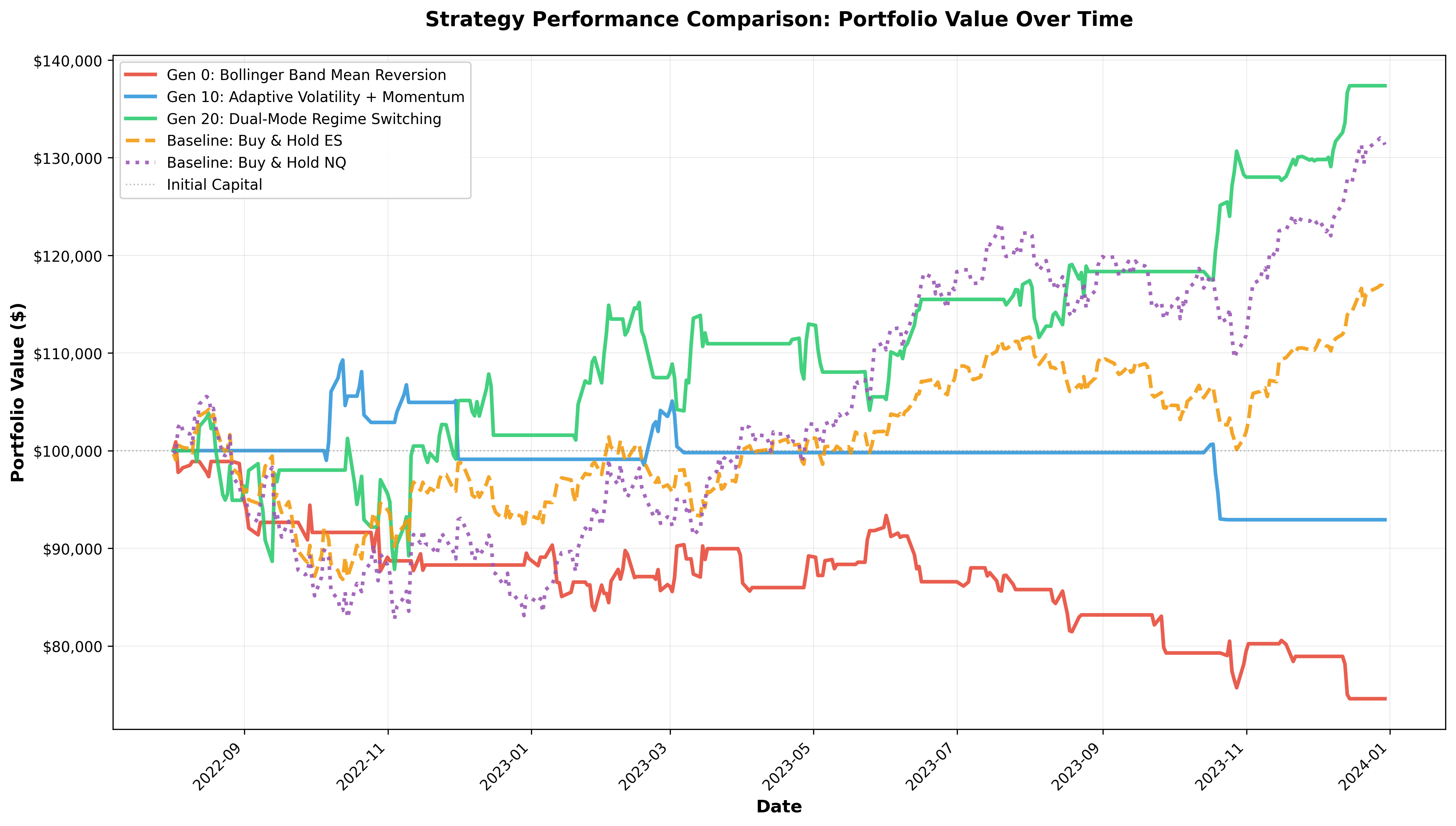}
    \caption{Portfolio value evolution for futures strategies and baselines over the test period.}
    \label{fig:future_strategies}
\end{figure}

To demonstrate QuantEvolve's scalability across asset classes, we applied the framework to ES and NQ futures contracts. \Cref{tab:performance_futures,fig:future_strategies} present the results.

\begin{table}[t]
    \centering
    \small
    \begin{tabular}{p{0.8cm}l|cccc}
        \toprule
        
        \multicolumn{2}{c}{\textbf{Model}}
        & \multicolumn{1}{c}{SR} & \multicolumn{1}{c}{MDD} & \multicolumn{1}{c}{IR} & \multicolumn{1}{c}{CR}
        \\
        
        \midrule[0.25pt]
        
        \textit{Baseline}
        & B\&H ES
        & 0.66 & -16.7\% & - & 14.4\%
        \\
        & B\&H NQ
        & 0.97 & -21.6\% & - & 31.3\%
        \\
        
        \midrule[0.25pt]
        
        \textit{\textbf{Ours}}
        & Gen 0
        & -1.21 & -26.1\% & -1.36 & -25.4\%
        \\
        & Gen 10
        & -0.41 & \textbf{-15.0\%} & -0.76 & -7.1\%
        \\
        & Gen 20
        & \textbf{1.03} & -15.4\% & \textbf{0.49} & \textbf{37.4\%}
        \\
        
        \bottomrule
    \end{tabular}
    \caption{
        Performance on futures test set (Aug 2022 - Jan 2024). Baselines represent individual buy-and-hold positions for ES and NQ contracts. Generation 20 outperforms both baselines across risk-adjusted metrics.
    }
    \label{tab:performance_futures}
\end{table}

The futures results demonstrate QuantEvolve's ability to generate effective strategies across asset classes with distinct characteristics. Generation 0's poor performance (SR=-1.21, CR=-25.4\%) confirms that naive mean-reversion fails in trending futures markets. The framework rapidly adapted: generation 10 substantially reduced drawdown to -15.0\%, while generation 20 achieved positive risk-adjusted returns (SR=1.03) that exceeded both ES (SR=0.66) and NQ (SR=0.97) baselines.

Notably, the evolved strategy delivered 37.4\% cumulative return compared to 14.4\% for ES and 31.3\% for NQ, with lower maximum drawdown than the NQ baseline (-15.4\% vs -21.6\%). The Information Ratio of 0.49 indicates consistent outperformance relative to the baseline, validating that the hypothesis-driven evolutionary process successfully identified regime-adaptive mechanisms suited to futures market dynamics. This cross-asset generalization—without manual recalibration of the framework—supports QuantEvolve's potential for scalable automated strategy discovery across diverse financial instruments.


\section{Ablation Study}
\subsection{Feature Bin Sizes}

\begin{figure}[t]
    \centering
    \includegraphics[width=0.95\columnwidth]{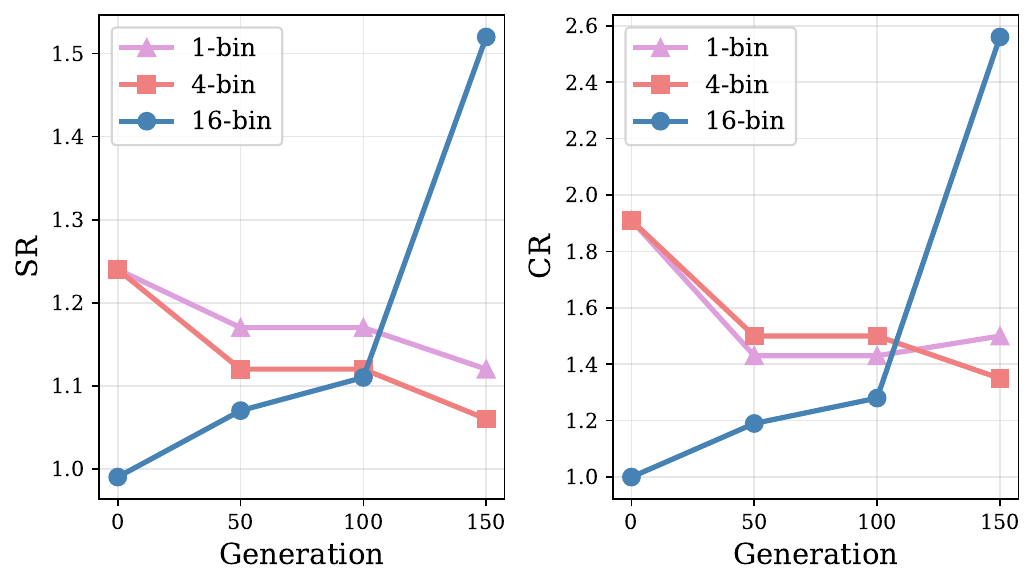}
    \caption{
        Impact of bin size on evolutionary performance. Higher bin resolution (16-bin) enables sustained improvement across generations, while lower resolutions (1-bin, 4-bin) lead to premature convergence.
    }
    \label{fig:ablation_bin_line}
\end{figure}

To investigate the impact of bin size on evolutionary dynamics, we conducted an ablation study comparing three configurations: 1-bin, 4-bin, and 16-bin per feature dimension.
The bin size determines the resolution of the feature map, directly influencing the balance between diversity maintenance and selection pressure.
Fewer bins reduce the number of distinct niches, potentially accelerating convergence but limiting exploration capacity, while more bins expand the behavioral space but may slow early-stage optimization.
In the extreme case of 1-bin, the feature map collapses to a single cell where all strategies compete directly, eliminating diversity maintenance.

\Cref{fig:ablation_bin_line} presents the evolution of test performance across generations for different bin size configurations. 
The 16-bin configuration exhibits sustained improvement throughout the evolutionary process, achieving a Sharpe ratio of 1.52 and a cumulative return of 256\% by generation 150.
This trajectory demonstrates continuous exploration and refinement, with performance gains accelerating in later generations as the framework identifies and exploits promising strategy patterns.

In contrast, both 1-bin and 4-bin configurations display early convergence followed by performance stagnation or degradation.
The 1-bin setting begins with relatively strong initial performance (SR = 1.24, CR = 191\%).
However, it fails to sustain improvement beyond generation 50, ultimately declining to SR = 1.12 and CR = 150\% by generation 150.
The 4-bin configuration follows a similar pattern, with initial performance of SR = 1.24 and CR = 191\% declining to SR = 1.06 and CR = 135\% by generation 150.

These results suggest that insufficient bin resolution leads to premature convergence by constraining niche diversity.
When the feature map contains too few cells, strategies with distinct behavioral characteristics are forced to compete within the same niche, eliminating potentially valuable exploration directions.
This competition favors strategies that perform well during training under the observed market conditions but may sacrifice the robustness and generalization capacity required for out-of-sample performance.
Furthermore, the sustained performance of the 16-bin configuration indicates that higher bin resolution enables the preservation of diverse strategy populations, potentially reducing the risk of overfitting.

\subsection{Feature Dimensions}

\begin{figure}[t]
    \centering
    \hfill
    \begin{subfigure}[b]{0.42\linewidth}
        \centering
        \includegraphics[trim={1.0cm 1.0cm 1.0cm 1.0cm}, clip, width=\linewidth]{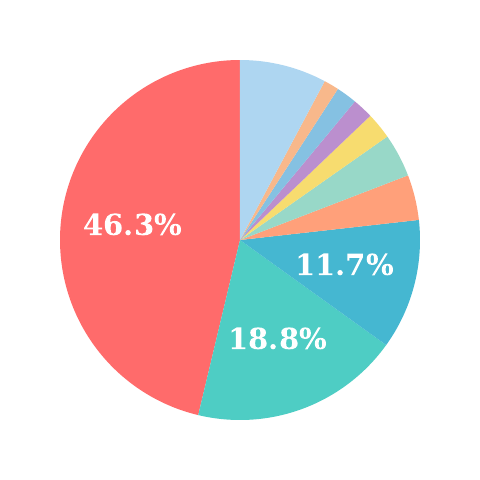}
        \subcaption{w/o category dimension}
        \label{fig:x_cat_43_category_pie}
    \end{subfigure}
    \hspace{10pt}
    \hfill
    \begin{subfigure}[b]{0.42\linewidth}
        \centering
        \includegraphics[trim={1.0cm 1.0cm 1.0cm 1.0cm}, clip, width=\linewidth]{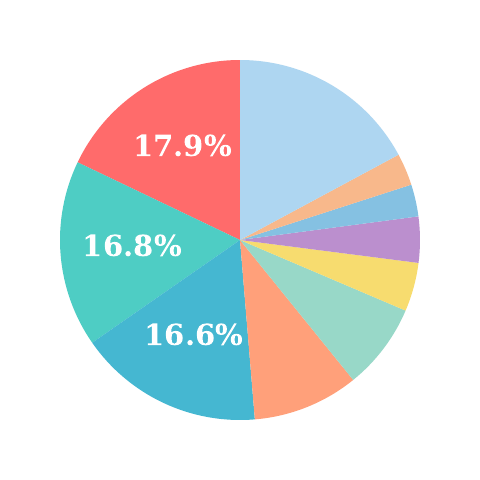}
        \subcaption{w/ category dimension}
        \label{fig:43_category_pie}
    \end{subfigure}
    \hfill
    \vspace{-5pt}
    \caption{
        Distribution of strategy categories across strategies in the evolutionary database. 
        Each pie slice represents the proportion of strategies in a given category, with the top 9 most frequent categories displayed individually and remaining categories aggregated.
        Percentages are shown for the top 3 categories only.
    }
    \label{fig:category_pie}
\end{figure}

To examine the impact of including the strategy category as a feature dimension, we compare two evolutionary runs: one excluding the category dimension (\Cref{fig:x_cat_43_category_pie}) and one including it (\Cref{fig:43_category_pie}).
Without the category dimension, the most dominant bin accounts for 46.3\% of all strategies, indicating a strong bias toward specific strategy categories.
This concentration suggests limited strategic diversity, with the evolutionary process favoring specific categories disproportionately.
Consequently, the framework may fail to adequately explore diverse strategic approaches, instead converging prematurely on a narrow subset of solution types.
In contrast, including the category dimension results in a more balanced distribution.
This more equitable allocation indicates that incorporating strategy category as a feature dimension effectively promotes population diversity and reduces the risk of premature convergence, enabling the evolutionary process to explore a broader range of strategic paradigms.


%% file: sec/appendix.tex
\newpage
\onecolumn
\newtcolorbox{promptbox}[1]{%
  enhanced, breakable,
  colback=gray!3, colframe=black, boxrule=0.8pt, sharp corners,
  title={#1}, coltitle=black,
  colbacktitle=orange!20,
  attach boxed title to top left={yshift=-\tcboxedtitleheight},
  boxed title style={colframe=black, colback=orange!20, boxrule=0.8pt, sharp corners},
  top=\tcboxedtitleheight+1mm, left=2mm, right=2mm, before skip=6pt, after skip=6pt
}

\newtcolorbox{redbox}[1]{%
  enhanced, breakable,
  colback=gray!3, colframe=black, boxrule=0.8pt, sharp corners,
  title={#1}, coltitle=black,
  colbacktitle=red!15,
  attach boxed title to top left={yshift=-\tcboxedtitleheight},
  boxed title style={colframe=black, colback=red!15, boxrule=0.8pt, sharp corners},
  top=\tcboxedtitleheight+1mm, left=2mm, right=2mm, before skip=6pt, after skip=6pt
}

\newtcolorbox{greenbox}[1]{%
  enhanced, breakable,
  colback=gray!3, colframe=black, boxrule=0.8pt, sharp corners,
  title={#1}, coltitle=black,
  colbacktitle=green!20,
  attach boxed title to top left={yshift=-\tcboxedtitleheight},
  boxed title style={colframe=black, colback=green!20, boxrule=0.8pt, sharp corners},
  top=\tcboxedtitleheight+1mm, left=2mm, right=2mm, before skip=6pt, after skip=6pt
}

\newtcolorbox{yellowbox}[1]{%
  enhanced, breakable,
  colback=gray!3, colframe=black, boxrule=0.8pt, sharp corners,
  title={#1}, coltitle=black,
  colbacktitle=yellow!20,
  attach boxed title to top left={yshift=-\tcboxedtitleheight},
  boxed title style={colframe=black, colback=yellow!20, boxrule=0.8pt, sharp corners},
  top=\tcboxedtitleheight+1mm, left=2mm, right=2mm, before skip=6pt, after skip=6pt
}

\newtcolorbox{graybox}[1]{%
  enhanced, breakable,
  colback=gray!3, colframe=black, boxrule=0.8pt, sharp corners,
  title={#1}, coltitle=black,
  colbacktitle=gray!20,
  attach boxed title to top left={yshift=-\tcboxedtitleheight},
  boxed title style={colframe=black, colback=gray!20, boxrule=0.8pt, sharp corners},
  top=\tcboxedtitleheight+1mm, left=2mm, right=2mm, before skip=6pt, after skip=6pt
}

\appendix

\section{Multi-agent Evolution Prompts}

\subsection*{Data Agent Prompts}

\begin{promptbox}{Data Agent – Feasible Categories (Island Creation) Prompt}
\begin{lstlisting}[breaklines=true]
You are a data agent specializing in quant trading strategy.
TASK: Generate a taxonomy of quantitative trading strategy families
that can be implemented using ONLY the provided data fields.
DATA STRUCTURE: {data_structure}
CONSTRAINTS:
- Use EXCLUSIVELY the fields listed in the data structure above
- Do NOT assume availability of: fundamentals, intraday tick data,
  options chains, alternative datasets, or external market data
- Output high-level strategy families only (no sub-strategies, indicators,
  or implementation details)
\end{lstlisting}
\end{promptbox}

\begin{promptbox}{Data Agent – Prompt to create Data Schema Prompt}
\begin{lstlisting}[breaklines=true]
# Based on the following data universe and its data structure, 
generate a CONSIDERATIONS section that explains the data structure 
and context parameters available for this specific universe.
Universe Structure Information:
{universe_structure}
{context_info}
Please return a message in the following style that starts with 
'DATA STRUCTURE:' and includes:
1. The actual data format (OHLCV, columns, sample data format)
2. The complete context parameters available in handle_data() 
3. Specific examples of how to access this universe data
4. Any bundle-specific considerations (futures vs equities, 
   daily vs minute data)
Follow this exact format and be specific to this bundle:
DATA STRUCTURE: [Describe the actual data format found in the CSV files]
Context Parameters Available in handle_data()
[Include the complete list of Zipline data context parameters, 
 with examples specific to this bundle]
Make it comprehensive and specific to the actual data structure 
found in this bundle.
ASSET CONSTRAINTS:
List only the base asset symbols available in this universe
\end{lstlisting}
\end{promptbox}

\subsection*{Strategy Agent Prompts}

\begin{redbox}{Strategy Agent – Hypothesis System Prompt}
\begin{lstlisting}[breaklines=true]
Your task is to create a new, novel, creative, high-performing trading strategy.
=====================================================================================
# Hypothesis test-driven approach:
Before writing the code, you must formulate a clear hypothesis about 
why this strategy will perform well. This hypothesis should be grounded 
in financial theory, market dynamics, statistical analysis, risk management, 
and historical patterns in stock markets. Make it as detailed and 
comprehensive as possible, including:
- The core hypothesis itself.
- The rationale: Why did you form this hypothesis? Reference specific 
  financial concepts, indicators, or data patterns.
- Objectives: What do you aim to test or achieve with this hypothesis 
  (e.g., improving Sharpe ratio, reducing drawdowns)?
- Expected insights: What can be learned if the hypothesis holds or fails? 
  How might it inform future iterations?
- Potential risks or limitations: Any edge cases, assumptions, or weaknesses 
  in the hypothesis.
- Next Step ideas: How this hypothesis could be varied or extended in 
  future iterations.
\end{lstlisting}
\end{redbox}

\begin{redbox}{Strategy Agent – Hypothesis User Prompt}
\begin{lstlisting}[breaklines=true]
# Structure this hypothesis and details using XML-like tags within a 
multi-line comment (using triple quotes """) at the very top of your output. 
For example:
    """
    <hypothesis> [Core hypothesis statement] </hypothesis>
    <rationale> [Detailed explanation of why this hypothesis was formed] </rationale>
    <objectives> [What you aim to test/achieve] </objectives>
    <expected_insights> [Potential learnings and implications] </expected_insights>
    <risks_limitations> [Any risks, assumptions, or weaknesses] </risks_limitations>
    <next_step_ideas> [Ideas for variations or future tests] </next_step_ideas>
    """
\end{lstlisting}
\end{redbox}

\subsection*{Code Team Prompts}

\begin{greenbox}{Code Team – Code Creation User Prompt}
\begin{lstlisting}[breaklines=true]
#Context: {context}
# Task
- Improve the performance of the current program on the specified metrics 
  by optimizing algorithms, reducing computational complexity, or enhancing 
  resource utilization.
- Fix issues in the current program by addressing bugs, resolving errors, 
  correcting logic flaws, or eliminating potential vulnerabilities that 
  affect functionality.
- Continue to validate the hypotheses of the current program through 
  comprehensive testing, verification of assumptions, and confirmation of 
  expected behaviors.
- Continue to experiment or analyze based on the analysis results of the 
  current program, building upon previous findings and insights.
- Improve the structure, readability, maintainability, scalability, 
  efficiency, or stability of the current program through code refactoring 
  and architectural enhancements.
- Try a completely new approach that addresses the same problem using 
  different methodologies, frameworks, or design patterns.
IMPORTANT: Make sure your rewritten program maintains the same inputs and 
outputs as the original program, but with improved internal implementation 
that enhances overall functionality and performance.
```{language}
# Your rewritten program goes here
\end{lstlisting}
\end{greenbox}

\begin{greenbox}{Code Team – Code Creation System Prompt}
\begin{lstlisting}[breaklines=true]
# System prompt used to generate the original code: {original_system_prompt}
# User prompt used to generate the original code: {original_user_prompt}
# Original code to be fixed:
```python
{original_code}
Error message received when running the original code: {error_message}
Traceback of the error: {traceback}
Please provide the corrected code:
\end{lstlisting}
\end{greenbox}

\subsection*{Evaluation Team Prompts}

\begin{yellowbox}{Evaluation Team – Evaluation System Prompt}
\begin{lstlisting}[breaklines=true]
You are an expert code reviewing team specializing in quantitative trading strategies.
Your job is to analyze the provided code and evaluate it systematically as a team.
# Important Constraints for Trading Strategy:
- The code must contain exactly two functions: `initialize(context)` and `handle_data(context, data)`.
- The `initialize` function sets up the trading strategy, defines equity, and initializes context variables.
- The `handle_data` function contains the trading logic and is called on each trading day.
- When implementing trading logic, you can only use past and present data.
- The strategy must avoid Lookahead Bias. If the strategy uses equity data for decision making, it is considered invalid. This is the most important constraint.
- You have to order less than or equal to the available cash.
- Properly handle exceptions and edge cases (e.g., insufficient data, contract rollover).
- The `initialize` function must include the following commission and slippage settings:
    `context.set_commission(commission.PerShare(cost=0.0075, min_trade_cost=1.0))`
    `context.set_slippage(slippage.VolumeShareSlippage())`
\end{lstlisting}
\end{yellowbox}

\begin{yellowbox}{Evaluation Team – Evaluation User Prompt}
\begin{lstlisting}[breaklines=true]
# Code to evaluate: ```python {current_program}```
# Code's performance metrics: {metrics}
# Additional information about the code's performance: {meta_info}
# At the end of your conversations and evaluations return the final answer as a JSON object with the following format:
{
    "hypothesis_evaluation_score": [float between 0.0 and 1.0],
    "hypothesis_evaluation_reasoning": [Detailed explanation],
    "program_alignment_evaluation_score": [float between 0.0 and 1.0],
    "program_alignment_evaluation_reasoning": [Detailed explanation, including sub-notes on readability, maintainability, efficiency],
    "results_analysis_score": [float between 0.0 and 1.0],
    "results_analysis_reasoning": [Detailed analysis, conclusions, and suggestions],
    "readability": [float between 0.0 and 1.0],
    "maintainability": [float between 0.0 and 1.0],
    "efficiency": [float between 0.0 and 1.0],
    "reasoning": "[Brief overall summary and recommendations for next iteration]",
    "insight": "[what ground truth does the result reveal. Synthesize the hypothesis and based on the result explain what insight we can infer  E.g. Asset X reacts Y way during Z]"
}
\end{lstlisting}
\end{yellowbox}

\begin{yellowbox}{Evaluation Team – Categorization System Prompt}
\begin{lstlisting}[breaklines=true]
You are a trading strategy categorizer. Your only task is to categorize the given trading strategy code within the provided list of strategy families.
**Strategy Families:**
{program_categories}
**Instructions:**
- Analyze the trading strategy hypothesis
- Select ONLY from the provided strategy families above (use ONLY Keys, NOT descriptions)
- If the strategy doesn't clearly fit any category, use "other-[category]" as a prefix
- Multiple categories are allowed if the strategy combines approaches
\end{lstlisting}
\end{yellowbox}

\begin{yellowbox}{Evaluation Team – Categorization User Prompt}
\begin{lstlisting}[breaklines=true]
You are an expert quantitative finance analyst specializing in trading strategy classification and taxonomy. Your role is to accurately categorize trading strategies based on their core logic and implementation approach.
# Key responsibilities:
- Analyze trading strategy hypotheses and implementations 
- Classify strategies according to established quantitative finance categories
- Focus on the fundamental trading logic rather than just technical indicators used
- Provide accurate, concise categorizations in the requested JSON format
- Do not create new categories or modify existing category names.
- Do not attempt to classify a strategy into all applicable categories at once. For each category within the given strategy families, carefully determine whether the given strategy truly belongs to that category on a one-by-one basis.
- Each categorization decision must be made after a thorough understanding of the category's definition and characteristics, as well as the given strategy's implementation and logic.
\end{lstlisting}
\end{yellowbox}

\section{Baseline Strategy Codes for Equity Markets}
\label{sec:baseline_code}






\begin{graybox}{Equal-Weighted Rebalancing}
\begin{lstlisting}[breaklines=true]
from zipline.finance import commission, slippage
from zipline.api import order_target_percent, record, symbols

def initialize(context):
    context.assets = symbols("AAPL", "NVDA", "AMZN", "GOOGL", "MSFT", "TSLA")
    context.set_commission(commission.PerShare(cost=0.0075, min_trade_cost=1.0))
    context.set_slippage(slippage.VolumeShareSlippage())

    # Equal weight for each asset
    context.target_weight = 1.0 / len(context.assets)

def handle_data(context, data):
    # Daily rebalancing to equal weights
    for asset in context.assets:
        if data.can_trade(asset):
            target_weight = context.target_weight

            order_target_percent(asset, target_weight)

    record(portfolio_value=context.portfolio.portfolio_value)

\end{lstlisting}
\end{graybox}

\begin{graybox}{MACD}
\begin{lstlisting}[breaklines=true]
from zipline.finance import commission, slippage
from zipline.api import order_target_percent, record, symbols

def calculate_macd(prices, fast=12, slow=26, signal=9):
    exp1 = prices.ewm(span=fast).mean()
    exp2 = prices.ewm(span=slow).mean()
    macd_line = exp1 - exp2
    signal_line = macd_line.ewm(span=signal).mean()
    histogram = macd_line - signal_line

    return macd_line, signal_line, histogram

def initialize(context):
    context.assets = symbols("AAPL", "NVDA", "AMZN", "GOOGL", "MSFT", "TSLA")
    context.set_commission(commission.PerShare(cost=0.0075, min_trade_cost=1.0))
    context.set_slippage(slippage.VolumeShareSlippage())

    # MACD parameters
    context.fast = 12
    context.slow = 26
    context.signal = 9
    context.lookback_window = max(context.slow + context.signal, 50)

def handle_data(context, data):
    try:
        # Get historical prices
        prices = data.history(
            context.assets, "close", context.lookback_window + 1, "1d"
        )

        # Equal weight if insufficient data
        if prices is None or prices.empty or len(prices) < context.lookback_window:
            equal_weight = 1.0 / len(context.assets)
            for asset in context.assets:
                if data.can_trade(asset):
                    order_target_percent(asset, equal_weight)
            record(portfolio_value=context.portfolio.portfolio_value)
            return

        # Calculate MACD for each asset
        macd_signals = {}
        for asset in context.assets:
            if asset in prices.columns:
                asset_prices = prices[asset].dropna()
                if len(asset_prices) >= context.lookback_window - 10:
                    macd_line, signal_line, histogram = calculate_macd(
                        asset_prices, context.fast, context.slow, context.signal
                    )

                    macd_signals[asset] = {
                        "signal": histogram.iloc[-1] > 0,
                    }

        # Calculate weights based on MACD signals
        buy_assets = [
            asset
            for asset, signal in macd_signals.items()
            if signal.get("signal", False)
        ]

        # Portfolio allocation logic
        if buy_assets:
            # Equal weight among buy signals
            target_weight = 1.0 / len(buy_assets)

            for asset in context.assets:
                if data.can_trade(asset):
                    weight = target_weight if asset in buy_assets else 0.0
                    order_target_percent(asset, weight)
        else:
            # No buy signals: equal weight cash-like allocation or stay in cash
            equal_weight = 1.0 / len(context.assets)
            for asset in context.assets:
                if data.can_trade(asset):
                    order_target_percent(asset, equal_weight)

    except Exception as e:
        equal_weight = 1.0 / len(context.assets)
        for asset in context.assets:
            if data.can_trade(asset):
                order_target_percent(asset, equal_weight)

    record(portfolio_value=context.portfolio.portfolio_value)

\end{lstlisting}
\end{graybox}

\begin{graybox}{RSI \& KDJ}
\begin{lstlisting}[breaklines=true]
from zipline.finance import commission, slippage
from zipline.api import order_target_percent, record, symbols
import pandas as pd

def calculate_rsi(prices, period=14):
    delta = prices.diff()
    gain = (delta.where(delta > 0, 0)).rolling(window=period).mean()
    loss = (-delta.where(delta < 0, 0)).rolling(window=period).mean()

    rs = gain / loss
    rsi = 100 - (100 / (1 + rs))
    return rsi

def calculate_stochastic(high, low, close, k_period=14, d_period=3):
    lowest_low = low.rolling(window=k_period).min()
    highest_high = high.rolling(window=k_period).max()

    k_percent = 100 * ((close - lowest_low) / (highest_high - lowest_low))
    d_percent = k_percent.rolling(window=d_period).mean()
    j_percent = 3 * k_percent - 2 * d_percent

    return k_percent, d_percent, j_percent

def initialize(context):
    context.assets = symbols("AAPL", "NVDA", "AMZN", "GOOGL", "MSFT", "TSLA")
    context.set_commission(commission.PerShare(cost=0.0075, min_trade_cost=1.0))
    context.set_slippage(slippage.VolumeShareSlippage())

    # Technical indicator parameters
    context.rsi_period = 14
    context.kdj_k_period = 14
    context.kdj_d_period = 3
    context.lookback_window = max(context.rsi_period, context.kdj_k_period) + 10

    # Signal thresholds
    context.rsi_oversold = 30
    context.rsi_overbought = 70
    context.kdj_oversold = 20
    context.kdj_overbought = 80

def handle_data(context, data):
    try:
        # Get historical price data
        prices = data.history(
            context.assets,
            ["close", "high", "low"],
            context.lookback_window + 1,
            "1d",
        )

        # Equal weight if insufficient data
        if prices is None or prices.empty or len(prices) < context.lookback_window:
            equal_weight = 1.0 / len(context.assets)
            for asset in context.assets:
                if data.can_trade(asset):
                    order_target_percent(asset, equal_weight)
            record(portfolio_value=context.portfolio.portfolio_value)
            return

        # Calculate technical indicators for each asset
        signals = {}
        for asset in context.assets:
            try:
                # MultiIndex DataFrame: index=(date, asset), columns=['close', 'high', 'low']
                # Use xs to cross-section by asset
                asset_data = prices.xs(asset, level=1)
                close_prices = asset_data["close"].dropna()
                high_prices = asset_data["high"].dropna()
                low_prices = asset_data["low"].dropna()

                if len(close_prices) >= context.lookback_window - 5:
                    # Calculate RSI
                    rsi = calculate_rsi(close_prices, context.rsi_period)
                    current_rsi = rsi.iloc[-1] if not pd.isna(rsi.iloc[-1]) else 50

                    # Calculate KDJ
                    k_pct, d_pct, j_pct = calculate_stochastic(
                        high_prices,
                        low_prices,
                        close_prices,
                        context.kdj_k_period,
                        context.kdj_d_period,
                    )
                    current_k = k_pct.iloc[-1] if not pd.isna(k_pct.iloc[-1]) else 50
                    current_d = d_pct.iloc[-1] if not pd.isna(d_pct.iloc[-1]) else 50
                    current_j = j_pct.iloc[-1] if not pd.isna(j_pct.iloc[-1]) else 50

                    # Generate signals
                    # Buy signal: RSI oversold AND KDJ oversold
                    rsi_oversold = current_rsi < context.rsi_oversold
                    kdj_oversold = (
                        current_k < context.kdj_oversold
                        and current_d < context.kdj_oversold
                    )
                    buy_signal = rsi_oversold and kdj_oversold

                    # Sell signal: RSI overbought OR KDJ overbought
                    rsi_overbought = current_rsi > context.rsi_overbought
                    kdj_overbought = (
                        current_k > context.kdj_overbought
                        or current_d > context.kdj_overbought
                    )
                    sell_signal = rsi_overbought or kdj_overbought

                    # Strong buy signal: Very oversold conditions
                    strong_buy = current_rsi < 25 and current_k < 15

                    # Signal strength (for position sizing)
                    if strong_buy:
                        signal_strength = 2.0
                    elif buy_signal:
                        signal_strength = 1.0
                    elif sell_signal:
                        signal_strength = 0.0
                    else:
                        signal_strength = 0.5

                    signals[asset] = {
                        "rsi": current_rsi,
                        "k": current_k,
                        "d": current_d,
                        "j": current_j,
                        "buy_signal": buy_signal,
                        "sell_signal": sell_signal,
                        "strong_buy": strong_buy,
                        "signal_strength": signal_strength,
                    }
            except (KeyError, IndexError, ValueError) as e:
                # Skip asset if data is not available
                continue

        # Portfolio allocation logic
        total_signal_strength = sum(
            signal.get("signal_strength", 1.0) for signal in signals.values()
        )

        if total_signal_strength > 0:
            for asset in context.assets:
                if data.can_trade(asset) and asset in signals:
                    signal_info = signals[asset]

                    # Weight based on signal strength
                    target_weight = (
                        signal_info.get("signal_strength", 1.0) / total_signal_strength
                    )

                    order_target_percent(asset, target_weight)
        else:
            for asset in context.assets:
                if data.can_trade(asset):
                    order_target_percent(asset, 0.0)

    except Exception as e:
        # Log error and fallback to equal weight
        equal_weight = 1.0 / len(context.assets)
        for asset in context.assets:
            if data.can_trade(asset):
                order_target_percent(asset, equal_weight)

    record(portfolio_value=context.portfolio.portfolio_value)

\end{lstlisting}
\end{graybox}

\begin{graybox}{Risk Parity}
\begin{lstlisting}[breaklines=true]
from zipline.finance import commission, slippage
from zipline.api import order_target_percent, record, symbols
import numpy as np

def initialize(context):
    context.assets = symbols("AAPL", "NVDA", "AMZN", "GOOGL", "MSFT", "TSLA")
    context.set_commission(commission.PerShare(cost=0.0075, min_trade_cost=1.0))
    context.set_slippage(slippage.VolumeShareSlippage())

    # Risk Parity parameters
    context.lookback_window = 60
    context.rebalance_freq = 1  # Rebalance every 1 days to reduce transaction costs
    context.day_count = 0

def handle_data(context, data):
    context.day_count += 1

    # Only rebalance every N days and after sufficient history
    if (
        context.day_count < context.lookback_window
        or context.day_count % context.rebalance_freq != 0
    ):
        return

    # Get historical prices for volatility calculation
    try:
        prices = data.history(
            context.assets, "close", context.lookback_window + 1, "1d"
        )

        if prices is None or prices.empty:
            record(portfolio_value=context.portfolio.portfolio_value)
            return

        # Calculate returns and volatility
        returns = prices.pct_change().dropna()

        if len(returns) < context.lookback_window:
            record(portfolio_value=context.portfolio.portfolio_value)
            return

        # Calculate rolling volatility (annualized)
        volatilities = returns.std() * np.sqrt(252)

        # Handle any NaN or zero volatilities
        volatilities = volatilities.fillna(volatilities.mean())
        volatilities = volatilities.replace(0, volatilities.mean())

        # Calculate inverse volatility weights
        inv_vol = 1 / volatilities
        weights = inv_vol / inv_vol.sum()

        # Rebalance portfolio
        for asset in context.assets:
            if data.can_trade(asset) and asset in weights.index:
                target_weight = weights[asset]

                order_target_percent(asset, target_weight)

    except Exception as e:
        equal_weight = 1.0 / len(context.assets)
        for asset in context.assets:
            if data.can_trade(asset):
                order_target_percent(asset, equal_weight)

    record(portfolio_value=context.portfolio.portfolio_value)

\end{lstlisting}
\end{graybox}

\begin{table}[t]
    \centering
        \begin{tabular}{ll}
            \toprule
            \textbf{Category} & \textbf{Description} \\
            \midrule
            Momentum/Trend & Momentum/Trend Following Strategies \\
            Mean-Reversion & Mean Reversion Strategies \\
            Volatility & Volatility-Based Strategies \\
            Volume/Liquidity & Volume/Liquidity Analysis Strategies \\
            Breakout/Pattern & Breakout/Pattern Recognition Strategies \\
            Correlation/Pairs & Correlation / Pairs / Cross-Asset / Statistical Arbitrage Trading Strategies \\
            Risk/Allocation & Risk Parity / Allocation / Risk Management Strategies \\
            Seasonal/Calendar Effects & Seasonal/Calendar Effects Strategies \\
            \bottomrule
    \end{tabular}
    \caption{Strategy categories in equity markets}
    \label{tab:strategy_category_equity}
\end{table}